\documentclass{article} 
\usepackage{iclr2027_conference,times}

\usepackage{fancyhdr} 
\usepackage{amsmath,amsfonts,bm}

\def\eqref#1{equation~\ref{#1}}

\def\1{\bm{1}}

\DeclareMathAlphabet{\mathsfit}{\encodingdefault}{\sfdefault}{m}{sl}
\SetMathAlphabet{\mathsfit}{bold}{\encodingdefault}{\sfdefault}{bx}{n}

\usepackage{wrapfig}
\usepackage{hyperref}
\usepackage{url}
\usepackage{graphicx}
\usepackage{subcaption}
\usepackage{pifont} 
\usepackage{amssymb}
\usepackage{booktabs}
\usepackage{multirow}

\newcommand{\rev}[1]{\textcolor{black}{#1}}

\usepackage[most]{tcolorbox} \newtcolorbox{takeawaybox}{ colback=blue!5, colframe=blue!60!black, title=Takeaway, fonttitle=\bfseries, arc=2mm }
\usepackage{placeins}
\usepackage{tikz}
\usetikzlibrary{arrows.meta,positioning,calc,fit,backgrounds}
\newtcolorbox{tobetested}{ colback=red!5, colframe=red!75!black, title=To Be Tested, fonttitle=\bfseries, breakable }
\newtcolorbox{conclusions}{ colback=green!5, colframe=green!75!black, title=Conclusions, fonttitle=\bfseries, breakable }

\title{When do stored facts become available in looped transformers?}

\author{%
\parbox{\textwidth}{\centering \vspace*{0.15in} Fabio Valerio Massoli\textsuperscript{1,*,$\dagger$} \qquad Arnau Padr\'es Masdemont\textsuperscript{1,*}\\ [0.1cm] Jordi Ros Giralt\textsuperscript{1} \qquad Arash Behboodi\textsuperscript{1} \\ [0.25cm] {\normalfont \textsuperscript{1}Qualcomm AI Research} } }

\iclrfinalcopy 

\begin{document}

\maketitle
\begingroup 
\renewcommand\thefootnote{*} \footnotetext{Equal contribution. \quad \textsuperscript{$\dagger$}Corresponding author: \texttt{fmassoli@qti.qualcomm.com}}
\renewcommand\thefootnote{1} \footnotetext{Qualcomm AI Research is an initiative of Qualcomm Technologies, Inc.} 
\endgroup

\begin{abstract}
Looped transformers do not acquire new knowledge during a forward pass. Prior work shows that they store as many bits in their weights as non-looped models. It remains unclear \emph{how recurrence makes this knowledge available during inference}. We study this question using synthetic factual recall and multi-hop reasoning, measuring depth-resolved readable information in bits with native readouts and tuned lenses. Under final-loop supervision, single-hop facts become readable progressively: heavier memorization loads delay retrieval despite comparable final readable information. Untied baselines of equal unrolled depth concentrate retrieval near the end of their stacks, whereas looped models expose facts earlier. This delay is not intrinsic to the tested recall tasks: training with supervision and depth-specific readouts at every loop makes near-final information readable after the first loop in all tested configurations. Representational analysis shows value-related structure emerging within states organized by the query key. We then study multi-hop chains whose facts are stored in the weights rather than given in the prompt. Intermediate entities become decodable in chain order, and aligned transitions show next-entity structure developing while current-entity structure remains dominant. Interventions in a released six-recurrence model establish a causal role for these representations. Counterfactual replacements of a probe-defined entity component at a single prompt position redirect downstream bridge entities and the final answer. Conversely, replacements from queries that reach the same entity through different preceding relations, preserving the remaining relation suffix, retain the original answer in over 99.5\% of tested cases. These results distinguish a training-dependent schedule for factual recall from the causal role of retrieved-entity representations in subsequent composition.

\end{abstract}

\begin{figure}[!ht]
\centering
\includegraphics[width=\linewidth]{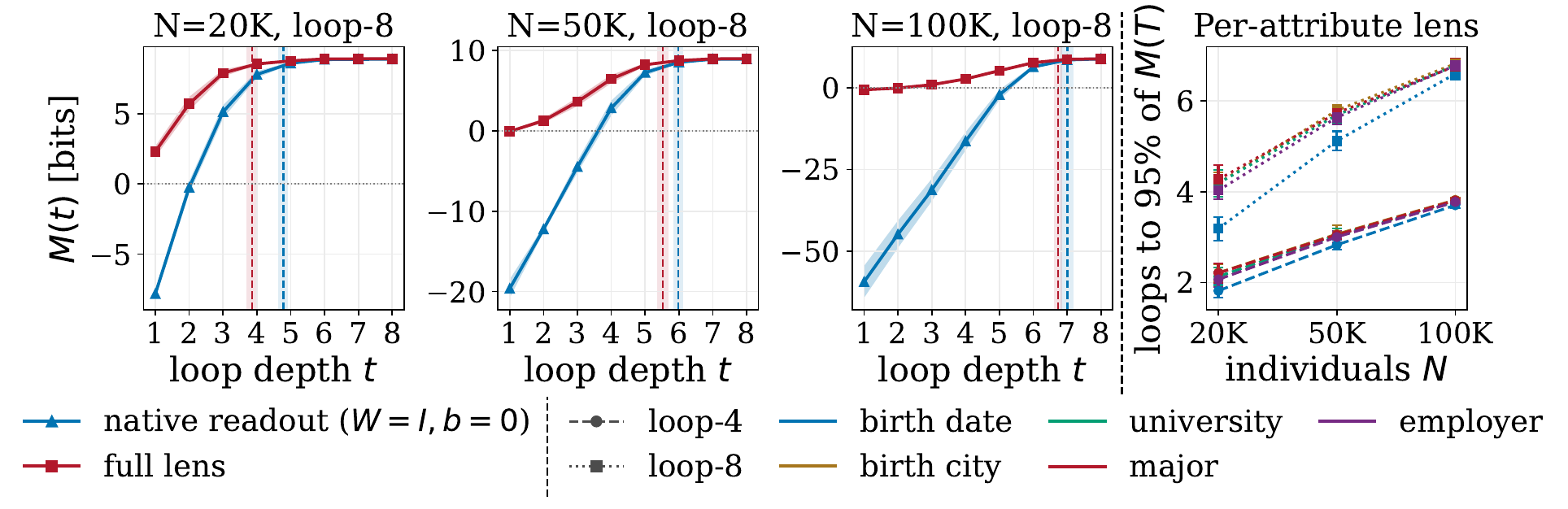}
\caption{\textbf{Readable bits $M(t)$ over five seeds (mean $\pm$1 SD).} \emph{Left}: native readout and full lens at loop-8; dashed lines and shaded spans mark the plateau depth. \emph{Right}: loops to 95\% of the final $M(t)$ per attribute against the number of individuals $N$ (full lens). 
}
\label{fig:native_vs_lens}
\end{figure}

\section{Introduction} \label{sec:intro}

Looped and non-looped transformers with the same number of parameters have comparable memorization capacity~\citep{zhu2025scaling}. Final-output measurements, however, do not reveal when stored facts become readable during inference. Nor does decoding information from an intermediate state establish how subsequent computation uses it. We investigate retrieval timing and causal influence through synthetic factual recall and multi-hop reasoning, asking two main questions:
\newline
\newline
\textbf{Q1.} \emph{When do stored facts become readable, and how does memorization load affect retrieval depth?}
\newline
\newline
\textbf{Q2.} \emph{Do retrieved intermediate-entity representations causally influence subsequent computation during multi-hop reasoning?}
\newline
\newline
For \textbf{Q1}, we measure readable information in bits at successive depths using native readouts and tuned lenses (Section~\ref{sec:singlehop}). Under final-loop supervision, single-hop facts become readable progressively: heavier memorization loads delay retrieval despite comparable final readable information. Equal-unrolled-depth untied baselines concentrate retrieval near the end of their stacks, whereas the tested looped models expose facts earlier. Representational analysis complements these readouts: structure associated with the queried individual (the key) is present from the first loop, while structure associated with the stored answer (the value) develops afterwards.
Moreover, we train matched variants with supervision and depth-specific readouts at every loop. In all eight tested configurations, near-final information becomes readable after the first recurrence (Appendix~\ref{app:every_loop}). The delayed retrieval observed under final-loop supervision is therefore not unavoidable in these recall settings: its timing depends on the training protocol.

For \textbf{Q2}, we turn to multi-hop reasoning, where each retrieved entity becomes the key for the next relation (Section~\ref{sec:multihop}). Concurrent work on prompt-provided graphs reports ordered intermediate-node decodability and answer steering through full-state interventions~\citep{sesodia2026insideloop}. We study composition of stored facts~\citep{ltg}: the prompt supplies the starting entity and relation sequence, but not the full graph or intermediate bridge entities. We recover the same ordered decodability in this setting. Our aligned-transition analysis further connects this progression to single-hop recall: next-entity structure develops while current-entity structure remains stronger across the analyzed transitions.

To test causal influence, we replace a probe-defined entity component of the hidden state at one token position. In the tested models, these localized interventions redirect downstream bridge entities within one further recurrence and change the final answer. We test specificity with random-subspace controls and with donors that reach the same entity through a different relation path. When the remaining relation suffix is matched, same-entity patches almost always preserve the original answer. Together, these results distinguish a training-dependent schedule for factual recall from the causal influence of intermediate-entity representations during composition.

Our contributions are:
\begin{itemize}
\item an entropy-referenced evaluation of factual readability, revealing load-dependent retrieval depth and earlier readability in looped models than in the tested equal-unrolled-depth untied baselines (Section~\ref{sec:setup}, Section~\ref{sec:schedule});
\item matched training comparisons showing that every-loop supervision with depth-specific readouts yields near-final first-loop readability in eight tested configurations (Appendix~\ref{app:every_loop});
\item a representational analysis connecting single-hop key/value structure to current-/next-entity structure during composition of stored facts (Section~\ref{sec:keyvalue}, Section~\ref{sec:multi_hop_retrieval});
\item evidence from localized entity-subspace interventions that intermediate representations causally influence downstream entities and final answers, with random-subspace and matched same-entity controls (Section~\ref{sec:causal_interventions}).
\end{itemize}

\section{Retrieving stored facts}
\label{sec:singlehop}

\begin{figure}[!ht]
\centering
\includegraphics[width=\linewidth]{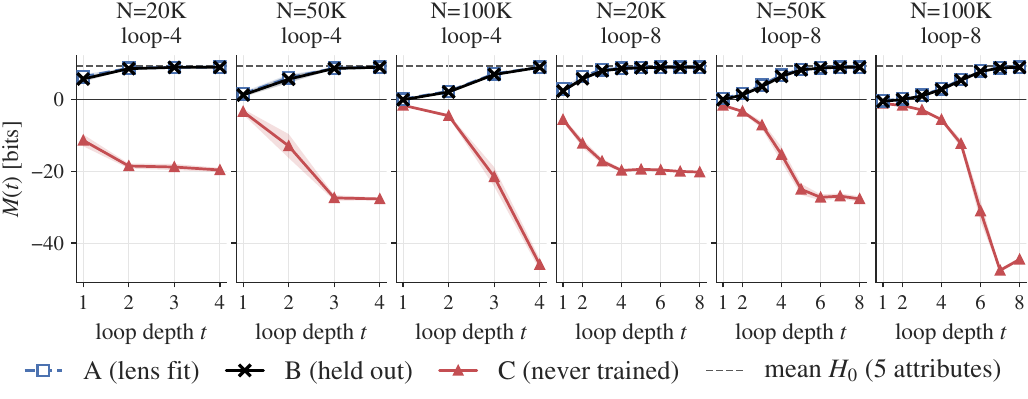}
\caption{\textbf{Lens controls.} Five-attribute mean $M(t)$ for the full lens on the Figure~\ref{fig:native_vs_lens} models (11-layer block, $d=192$). We report the mean $\pm$1 SD over five seeds. Dashed horizontal lines report $H_0$.}
\label{fig:lens_controls}
\end{figure}

We study when memorized facts become readable, how this timing changes with load, and how individual- and answer-associated structure develop across recurrences.

\subsection{Experimental setup} \label{sec:setup}

\paragraph{Models and data.} Following Ouro \citep{zhu2025scaling}, we train weight-tied, GPT-2-style Transformers with an 11-layer shared block, width $d=192$, and loop budgets $L\in\{4,8\}$. BioS \citep{allen2023physics} supplies synthetic biographies of $N\in\{20\text{K},50\text{K},100\text{K}\}$ individuals; we score birth city, university, employer, major and birth date (Appendix~\ref{app:training}). Individuals are split into three sets, namely, A/B/C ($60/35/5\%$): the backbone trains on A$\cup$B, read-out lenses on A only, and C is never used in any training. We use final-loop-only loss and five training seeds per $(N,L)$ tuple. Attribute values are sampled independently of names, so the lens cannot generalize by learning a name--attribute rule shared across A and B. The A/B comparison instead tests whether a decoder learned on one set of backbone-trained individuals transfers to another; C tests the decoder on individuals absent from training. We report in Appendix~\ref{app:lens_controls} results for every-loop loss.

\paragraph{Readable bits.} For an attribute sampled uniformly from $K$ values, the generating source has entropy $H_0=\log_2 K$ (e.g.\ $7.64$ bits for 200 cities). We compute the readable bits, at depth $t$, as: $M(t) = H_0-\mathrm{CE}(t)$ \citep{barber2004algorithm}. Thus, $M(t)=0$ matches a uniform-value log-loss baseline, $M<0$ is worse than that baseline, and $M(t)=H_0$ means zero cross-entropy. 
We interpret $M$ as the reduction in prediction loss, measured in bits, relative to guessing uniformly over the generator's possible attribute values.

\paragraph{Reading intermediate depths.} The native readout is the terminal LayerNorm and embedding-tied LM head. We also train a tuned lens \citep{belrose2023eliciting} at each depth:
$z^{(t)}=\texttt{lm\_head}\big(\texttt{ln}_f(W_t h^{(t)}+b_t)\big)$. Only the lens parameters $W_t$ and $b_t$ are trained; the backbone, final LayerNorm and output head remain frozen. We train the lenses by minimizing cross-entropy on the correct attribute tokens. The native readout applies the final LayerNorm and output head directly to $h^{(t)}$; the tuned lens applies them to $W_t h^{(t)}+b_t$. Both readouts therefore evaluate the same hidden state, but differ in how they decode it. Their performance gap does not directly measure information missing from that state. We report both readouts. 

\begin{figure}[!ht]
\centering
\includegraphics[width=\linewidth]{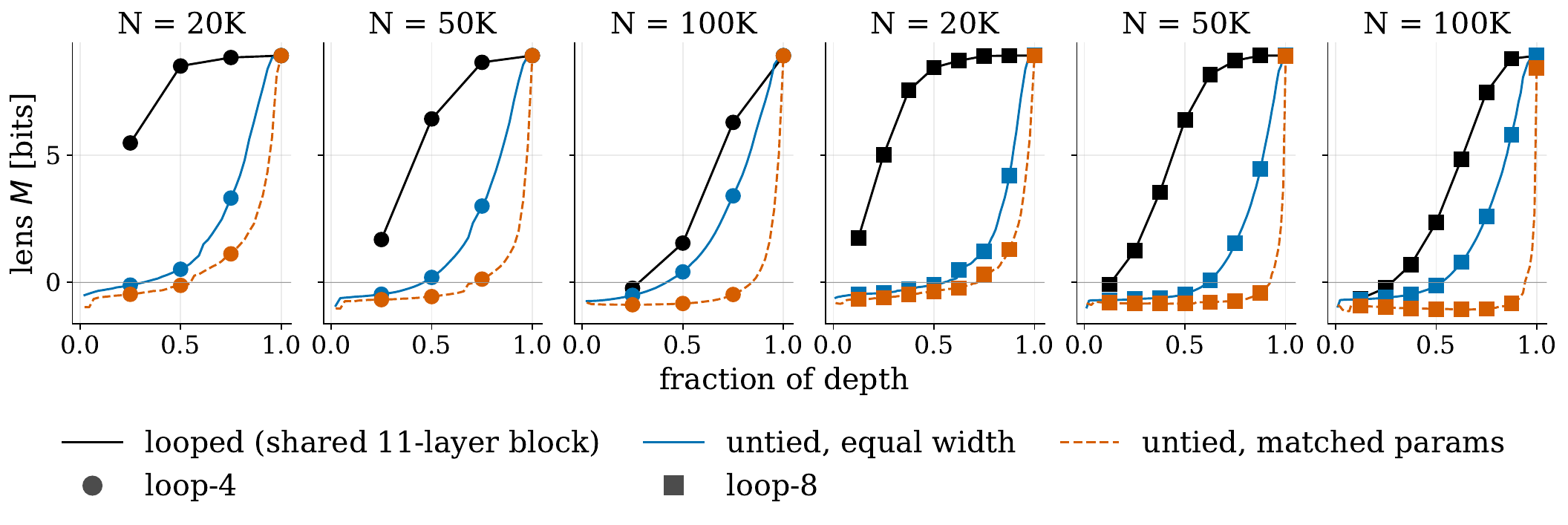}
\caption{\textbf{Looped versus untied models of equal unrolled depth.} Full-lens $M$ on B. Left: 44 layers; right: 88. Black: looped; blue: untied, equal width; orange: untied, approximately parameter-matched. Untied markers mark loop boundaries. See Appendix~\ref{app:untied} for sampling details.}
\label{fig:untied}
\end{figure}

\subsection{More stored facts use deeper retrieval} \label{sec:schedule}

\paragraph{Facts become readable progressively.} Figure~\ref{fig:native_vs_lens} (Left) shows that mean $M(t)$ across the five attributes increases with recurrence depth in loop-8 models. Final-loop lens scores reach approximately $94.5$--$97.5\%$ of the corresponding source entropy. Since the backbone weights shared, these profiles characterize changes in the readability of stored facts during looping. Tuned lenses yield higher scores than the native readout at intermediate depths, but their scores increase across recurrences as well. Thus, fitting a lens at each depth does not eliminate the observed depth dependence. Appendix~\ref{app:extra} reports the profiles for individual attributes and extends the analysis to loop-4 models

\paragraph{Depth scales with load.} Higher loads shift readability to later loops, while final lens scores remain nearly unchanged. For each attribute, seed-averaged $M(L)$ differs by less than $0.007$ bits across loads and loop budgets, with seed standard deviations below $0.01$ bits. We define completion depth as the first depth at which the five-attribute mean reaches $95\%$ of its final value, using linear interpolation between successive loops without smoothing or enforcing monotonicity. This threshold is relative to each curve's final prediction score, rather than to source entropy or accuracy, and does not require the curve to plateau. At $N=20\text{K},50\text{K},100\text{K}$, completion occurs at $0.50\%$, $0.74\%$, $0.94\%$ of the 4-loop budget and $0.48\%$, $0.69\%$, $0.84\%$ of the 8-loop budget, respectively (means over five seeds). Figure~\ref{fig:native_vs_lens} (Right) reports completion depths for individual attributes.
On average, the 4-loop model reaches the threshold near the end of its budget at the highest load. 
Across the three loads, the 8-loop model requires $1.78-1.94$ times as many recurrences as the 
4-loop model to reach the same relative threshold.

\paragraph{Lens generalization and capacity.} We evaluate whether the lenses generalize beyond their training individuals and whether the depth profiles depend on lens capacity. Across all models, the five-attribute mean scores on A and B differ by less than $0.71$ bits at every measured depth, whereas C scores remain below the uniform-value baseline for every attribute (Figure~\ref{fig:lens_controls}). To test sensitivity to lens capacity, we restrict the learned matrix to an identity map plus a low-rank correction, $W_t=I+U_tV_t$, testing correction ranks of 16, 32 and 64. The rank constraint applies only to $U_tV_t$; the identity term preserves access to the full hidden state. Figure~\ref{fig:ranks} in Appendix~\ref{app:ranks} compares these three correction ranks with the unrestricted lens; Appendix~\ref{app:lens_controls} provides the fitting protocol and additional decoder controls. The rank-constrained lenses retain the qualitative delay in readability with increasing load, although early-depth scores and some completion depths change. Thus, the load-dependent trend persists across the tested lens capacities, while score magnitudes and completion depths remain sensitive to the readout.

\paragraph{Single-pass comparisons.} 
We compare the looped models with parameter-matched, single-pass, non-looped models using lenses fitted at each layer. At $N=20\text{K},50\text{K},100\text{K}$, the five-attribute mean reaches $50\%$ of its final value at $65\%$, $83\%$, $92\%$ of the single-pass models' depth, compared with $25\%$, $40\%$, $65\%$ for loop-4 and $23\%$, $42\%$, $61\%$ for loop-8, respectively (Appendix~\ref{app:ranks}). All depths are normalized by the corresponding model's total unrolled depth. Threshold crossings are linearly interpolated between layer outputs for single-pass models and between loop outputs for looped models. The $92\%$ estimate falls between the final two layer outputs. For loop-4 at 20K, the threshold is already met at the first recurrence, so $25\%$ is an upper bound on the crossing depth. Then, we compare the looped with untied models of the same unrolled depth (44 or 88 layers). The untied models reach the same $50\%$ criterion at $81\%$--$88\%$ of their depth when width is held fixed, and at $93\%$--$99\%$ when width is reduced to approximately match the non-embedding parameter count (Figure~\ref{fig:untied}; Table~\ref{tab:untied} in Appendix~\ref{app:untied}). Final lens scores are comparable to those of the looped models, except for the reduced-width, 88-layer model at 100K. Previous work has documented layer-wise factual recall in standard Transformers \citep{geva2023dissecting,biran2024hopping}. In our comparisons, the looped models reach half of their final lens score at a smaller fraction of their total unrolled depth than either class of single-pass model.

\textbf{Takeaway:} \emph{
For \textbf{Q1}, stored facts become progressively readable under final-loop supervision; higher loads delay near-final readability despite comparable final scores. Evaluations should therefore report depth-resolved readability across loads, rather than final scores alone. The delay can be reduced: every-loop supervision combined with depth-specific readouts yields near-final first-loop readability in all eight tested configurations (Appendix~\ref{app:every_loop}). 
}

\subsection{Key and value structure across loops}\label{sec:keyvalue}

Beyond prediction quality, we examine whether hidden states distinguish individuals or attribute values. We introduce the following terminology. We define \emph{Key structure} the differences between individuals relative to variation across paraphrases of the same individual, and \emph{Value structure} the differences between groups with different correct attribute values, relative to variation among states sharing the same value.

\begin{wrapfigure}{r}{0.5\textwidth}
\vspace{-5mm}
\centering
\includegraphics[width=\linewidth]{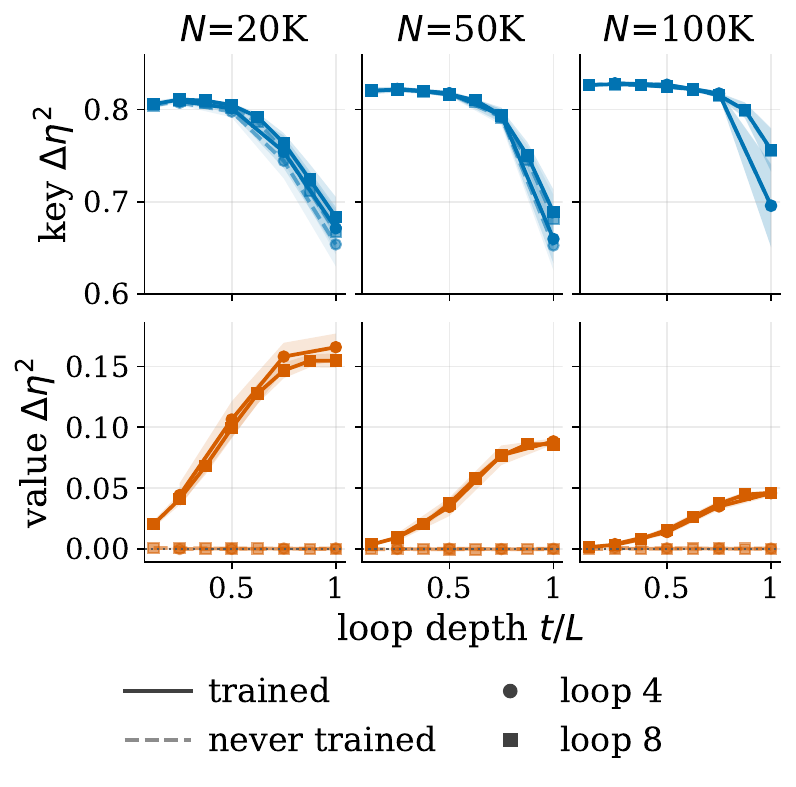}
\caption{\textbf{Key (top) and value (bottom) structure.} Null-subtracted explained variance, $\Delta\eta^2$; four-attribute means $\pm$1 SD over five seeds. Solid: A; dashed: never-trained C.}
\label{fig:key-value}
\vspace{-5mm}
\end{wrapfigure}

\paragraph{Key and value scores.} 
For each of the A and C splits, we sample $P=1000$ individuals with $Q=6$ paraphrases each. At every recurrence, we collect residual states at the prediction positions for the four single-token attributes: birth city, university, major and employer. For each split and attribute, we group the same $n=PQ$ states by individual or by value. For either grouping $\mathcal G$, the following ratio compares variation between group means with variation within groups:
\begin{equation} 
F(\mathcal{G};t)= \frac{\frac{1}{G-1}\sum_g n_g\lVert\bar h_g-\bar h\rVert_2^2} {\frac{1}{n-G}\sum_g\sum_{i\in g}\lVert h_i-\bar h_g\rVert_2^2}, \label{eq:fratio} 
\end{equation}
Here, $G$ is the number of groups, $n_g$ their sizes, and $\bar h_g$ and $\bar h$ the group and overall means at depth $t$. The corresponding fraction of explained variance is $\eta^2=F(G-1)/[F(G-1)+n-G]$.
We compare each score with a randomized reference that preserves group sizes. For keys, we randomly reassign individual states across groups. For values, we reassign individuals across value groups, keeping each individual's six paraphrases together. This preserves within-individual similarity, so the reference accounts for identity structure unrelated to the correct value. We repeat each procedure and average its $F$ scores to obtain $\overline{F}_{\mathrm{perm}}$, then define
\begin{equation}
\eta^2_{\mathrm{null}} = \frac{(G-1)\overline{F}_{\mathrm{perm}}}{(G-1)\overline{F}_{\mathrm{perm}}+n-G}.  \label{eq:eta} 
\end{equation}
In Figure~\ref{fig:key-value} we report $\Delta\eta^2=\eta^2-\eta^2_{\mathrm{null}}$: positive scores indicate more explained variance than the corresponding randomized reference; scores near zero indicate comparable explained variance.

\paragraph{Value structure across depths and loads.}
For backbone-trained individuals in A, key scores decrease from $\Delta\eta^2_{\rm key}=0.81$--$0.83$ at the first recurrence to $0.66$--$0.76$ at the last, while value scores increase to $0.15$--$0.17$, $0.09$ and $0.05$ at 20K, 50K and 100K, respectively (average over five seeds; Figure~\ref{fig:key-value}). Key scores remain larger throughout. Although grouping by individual necessarily explains at least as much unadjusted variance as grouping by value, this ordering is not guaranteed after subtracting the separate permutation baselines. The unseen individuals in C have key scores within $0.02$ of A's at every depth, but value scores remain near the randomized reference ($|\Delta\eta^2|<0.01$). Thus, separation by identity alone does not establish memorization; separation by the correct attribute value distinguishes trained from unseen individuals. To quantify its timing, we define $R(t)$ as the four-attribute mean of $F_{\rm value}(t)/F_{\rm perm}(t)$ on A, where $F_{\rm perm}$ is the mean score under value-group permutations. The midpoint $t_{1/2}$ is the first depth at which $R(t)$ reaches $1+[R(L)-1]/2$, halfway between the randomized baseline of one and its final value. This criterion uses the normalized $F$ ratio, not the plotted $\Delta\eta^2$; crossings are interpolated linearly in $\log[R(t)-1]$ between positive bracketing values (Appendix~\ref{app:keyvalue_timing}). At 20K, 50K and 100K, $t_{1/2}/L$ is $0.50$, $0.64$, $0.68$ for loop-4 and $0.49$, $0.59$, $0.62$ for loop-8, respectively (five-seed means; SD $0.02$--$0.04$; Figure~\ref{fig:seeds_thalf}). Higher loads therefore delay this relative midpoint and reduce the final value-associated explained variance. 


\section{Composing Stored Facts} \label{sec:multihop}

To address \textbf{Q2}, we trace intermediate entities during multi-hop retrieval, examine how a retrieved value becomes the key for the next hop, and test whether these representations causally influence subsequent computation.

\subsection{Experimental setup} \label{sec:setup_multihop}
\begin{figure}[t]
\centering
\includegraphics[width=\linewidth]{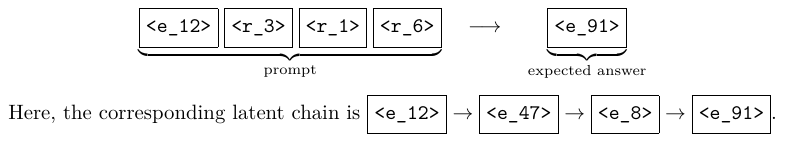}
\caption{\textbf{Composing stored atomic facts.} A 3-hop test query supplies the starting entity and relations, but not the intermediate steps.}
\label{fig:mh_prompt}
\end{figure}

\paragraph{Task and models.}  We study the fixed-depth (R6) and variable-depth (R\_dyn) looped transformers of \citet{ltg}, using their released checkpoints and three additional training seeds per regime trained with their code, data, and curriculum. Both use a four-layer recurrent block of width 768 and a vocabulary of 217 entity, relation and special tokens. A $k$-hop query $(e_0,r_1,\ldots,r_k)$ specifies the chain $e_0 \xrightarrow{r_1} e_1 \xrightarrow{r_2} \cdots \xrightarrow{r_k} e_k$ (Figure~\ref{fig:mh_prompt}). Only the endpoint $e_k$ is supervised as the answer; intermediate steps are neither supplied in the query nor included in its target.

\paragraph{Intermediate-entity probes.} 
We fit linear classifiers to hidden states concatenated across valid chain positions, separately for each loop, physical layer and hop index. Each loop--layer pair is one step of effective depth. Per-token probes localize these signals; Appendix~\ref{app:fur_mh} provides fitting details and comparisons with other readouts.

\subsection{Information retrieval in multi-hop chains}
\label{sec:multi_hop_retrieval}

\begin{figure}[!ht]
      \centering
    \includegraphics[width=\textwidth]{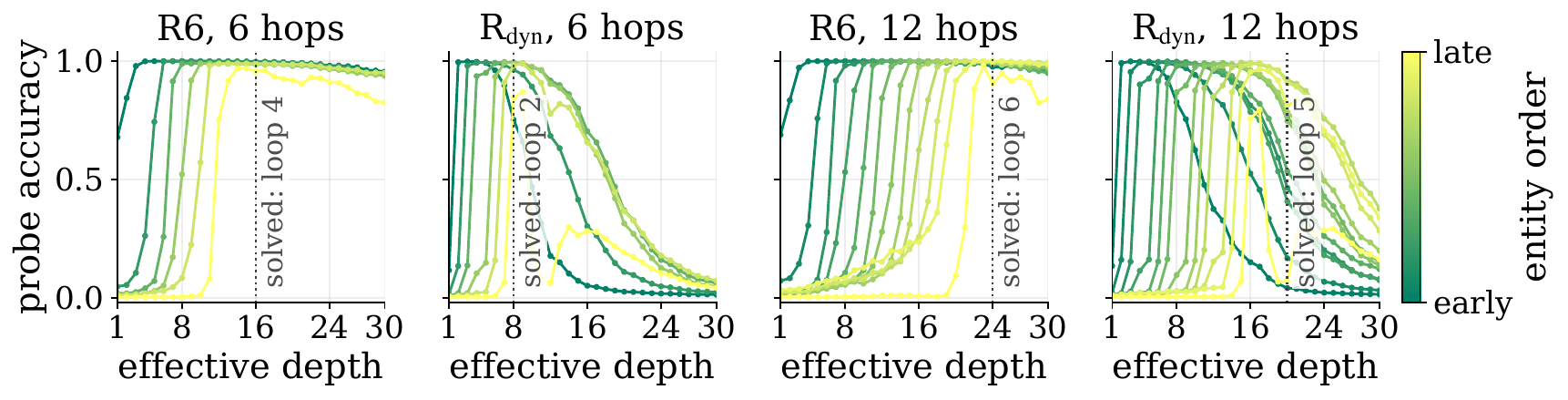}
      \caption{\textbf{Ordered intermediate-entity decoding in the released checkpoints.} Probe accuracy for 6-hop and 12-hop chains against effective depth (one step per loop--layer pair). Colors indicate entity order; dotted lines mark the first loop reaching 80\% endpoint accuracy. Probes cover 10 loops; the axis ends at effective depth 30.}
      \label{fig:multihop_r6_vs_rd}
  \end{figure}

\paragraph{Entities emerge in logical order.}
In both checkpoints \citep{ltg}, successive bridge entities become decodable as recurrent depth increases (Figure~\ref{fig:multihop_r6_vs_rd}). This complements the endpoint-accuracy staircase of \citet{ltg} and parallels concurrent probe results for in-context graph traversal \citep{sesodia2026insideloop}, here with each hop retrieved from facts stored in the weights.

\paragraph{State turnover and reasoning depth.} 
Earlier entities remain linearly decodable in R6 after later entities appear, whereas their decodability declines in R\_dyn. This turnover accompanies earlier endpoint prediction: R\_dyn reaches 80\% accuracy at loops 2 and 5 for 6- and 12-hop chains, versus loops 4 and 6 for R6. 

\begin{wrapfigure}{r}{0.5\textwidth}
\vspace{0mm}\centering
\includegraphics[width=\linewidth]{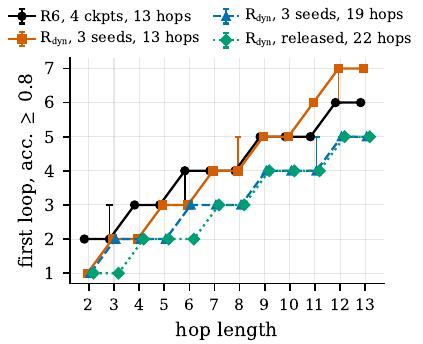}
\caption{\rev{\textbf{First loop reaching 80\% endpoint accuracy}, by hop length (750 test chains each). Markers: median across checkpoints; bars: min--max. R6: three retrained seeds and the released model. R\_dyn: three seeds at the 13- and 19-hop curriculum stages, plus the released 22-hop checkpoint (Appendix~\ref{app:mh_seeds}).}}
\label{fig:mh_seed_first_loop}
\vspace{0mm}
\end{wrapfigure}

\paragraph{Fixed versus dynamically sampled loop counts.} Before answering \textbf{Q2}, we found it relevant to assess the benefit of using sampled depths compared to fixed depth during training. Therefore, we asked the following question:

\emph{Does training with dynamically sampled loop counts lead to earlier answers than training with a fixed loop count?} 

We compare the two training regimes of \citet{ltg}: six loops in every batch versus a sampled depth per batch. The released checkpoints also differ in maximum training-chain length (13 versus 22 hops), so we train three seeds per regime and compare them after completing the same 13-hop curriculum stage. We measure the first inference loop reaching 80\% endpoint accuracy at each chain length (Figure~\ref{fig:mh_seed_first_loop}). Relative to fixed-loop training, dynamically sampled training reaches this threshold at most one loop earlier on 2--6-hop chains, but up to one loop later on 11--13-hop chains. In particular, all three dynamically trained seeds require seven loops for 13-hop chains, compared with six under fixed-loop training. Continuing those same seeds to the 19-hop stage yields earlier solutions, but changes both training-chain length and training duration. Thus, the earlier endpoint prediction observed in the released dynamically trained checkpoint is not a uniform advantage at a matched curriculum stage. 
We report additional details in Appendix~\ref{app:mh_seeds}.


\subsection{Causal patches on reasoning chains}
\label{sec:causal_interventions}

To distinguish decodability from causal influence, we replace part of a frontier state and test whether downstream entities and the final answer follow the counterfactual chain.

\begin{figure}[t]
  \centering
  \includegraphics[width=\linewidth]{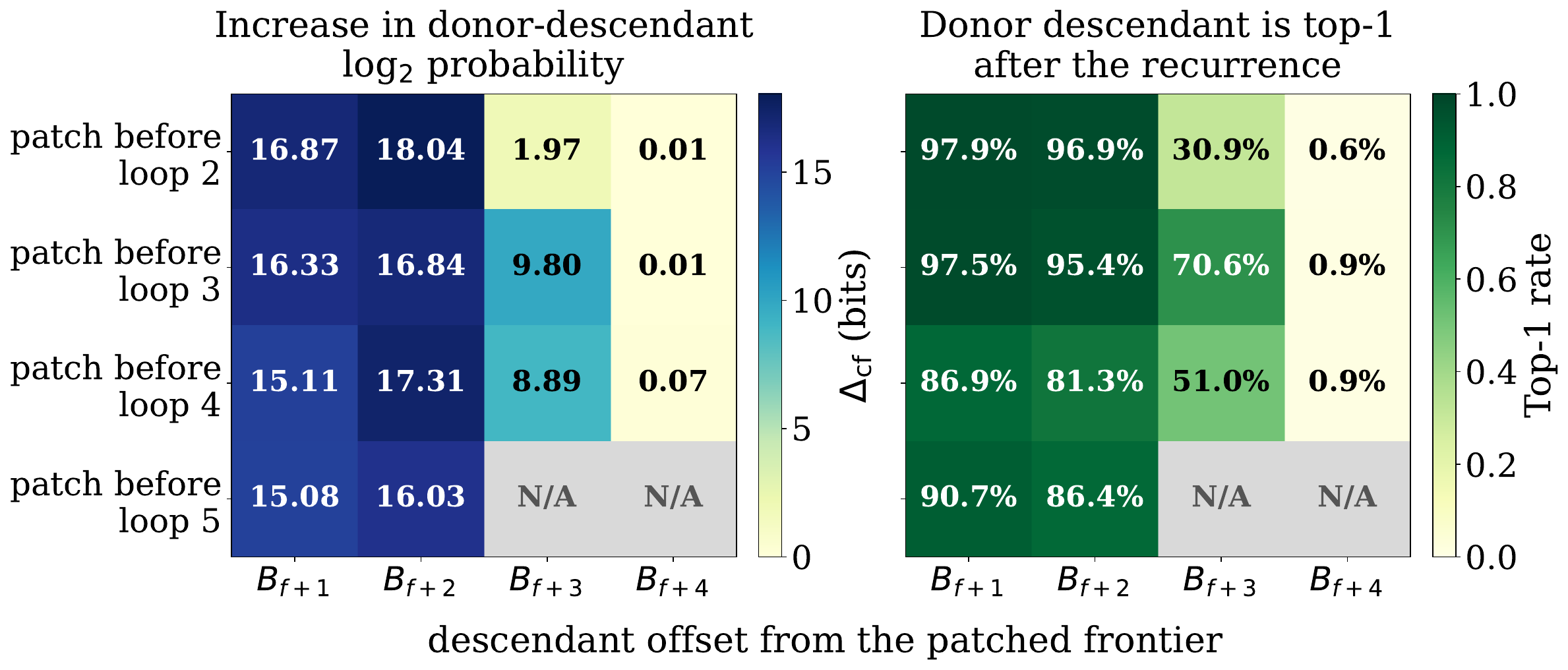}
  \caption{\textbf{One-recurrence causal reach after frontier patching.} Patches redirect nearby descendants, with reach varying across intervention depths. Three evaluation seeds pooled; 95\% bootstrap CIs over relation groups (Appendix~\ref{app:further_counterfactual}).}
  \label{fig:frontier-reach}
\end{figure}

\paragraph{Where to intervene.} Per-token probes decode a bridge entity $e_j$ at the position of $r_j$, the relation that produces it. We patch that position in a \emph{base} chain using a state from a \emph{donor} chain with a different starting entity but the same remaining relation suffix.

\paragraph{\rev{Controlled frontier patching.}}
We test whether changing a decoded intermediate representation redirects the remaining computation, using the released fixed-loop checkpoint of \citet{ltg} on held-out ten-hop chains with six inference loops. At the start of each loop, we identify the \emph{frontier}: the last entity in the chain prefix for which every entity has been correctly decoded with high confidence. We select the most common frontier position on validation data and retain base--donor pairs that both reach that position, matching them by how far their chains are decodable rather than by loop count alone. 
At the frontier's token position $p$, we replace only the component of the base state along entity-sensitive directions identified by the probe: $h'_p=h_p+P_{\mathrm{entity}}(h^{\mathrm{donor}}_p-h_p)$, where $P_{\mathrm{entity}}$ projects onto those directions. This transfers the donor's component in that subspace while leaving the orthogonal component and all other token states unchanged. We then resume the model and ask whether downstream entities and the final answer follow the donor-implied continuation. 
Three evaluation seeds vary the sampled relation groups and donors, not the model weights. Appendix~\ref{app:further_counterfactual} gives the confidence criterion, projector construction and controls; Table~\ref{tab:patch_counts} reports the selected frontiers and pair counts.

\paragraph{Patches propagate to downstream entities.} For interventions at recurrences 2--4, the frozen probes read each donor-implied descendant after one further recurrence. Descendants at offsets one and two gain 15.1--18.0 bits of probe log-probability relative to the unpatched states and are decoded top-1 in 81.3--97.9\% of cases (Figure~\ref{fig:frontier-reach}). The third gains 2.0--9.8 bits and is top-1 in 30.9--70.6\% of cases; the fourth is essentially unaffected. These per-descendant results show that one application of the four-layer block can propagate an intervention across multiple hops, with reach depending on the recurrent state.

\paragraph{\rev{Patches steer the final answer.}}\rev{After all six loops, entity-subspace patches at recurrences 1--5 yield the donor-implied endpoint in 79.07--98.57\% of cases (95\% bootstrap CIs within $\pm2.3$ percentage points; Figure~\ref{fig:frontier-answer}). At recurrences 2--5, they exceed equal-rank, norm-matched random-subspace patches by 67.5--81.6 points; at recurrence 1, the random control already reaches 50.95\%. For interventions at recurrences 1--4, the counterfactual endpoint is rarely predicted before output loop 5 (at most 0.4\%), but reaches 77.3--97.9\% at loop 5. Together with the downstream probe results, this supports propagation through subsequent computation rather than immediate endpoint readout.
}

\paragraph{Same-entity context control.} Donors that reach the same frontier entity through a different relation prefix, with the remaining suffix held fixed, preserve the original answer in 99.57--99.87\% of cases under entity-subspace patching and 98.33--99.52\% under whole-state replacement (recurrences 2--5; Appendix~\ref{app:further_counterfactual}). This extends evidence for functionally interchangeable bridge states across retrieval contexts from two-hop reasoning \citep{zhang2026knowinghopsenoughunderstanding} to these ten-hop chains.

\textbf{Takeaway:} \emph{
Answering \textbf{Q2}, intermediate-entity representations causally influence subsequent reasoning in the tested R6 checkpoint on ten-hop chains. Patching a probe-defined subspace at one relation position redirects downstream entities and yields the donor-implied endpoint in 79--97\% of cases at recurrences 2--5, whereas same-entity donors largely preserve the answer across retrieval contexts. The current bridge's state at its relation position is therefore a localized target for inspecting and steering this computation.
}

\section{Related work}

\paragraph{Memorization capacity.} \citet{allen2023physics} introduce synthetic biographies, and \citet{allen2023physics} report an empirical capacity of about two bits per parameter under sufficient training. \citet{zhu2025scaling} show that looping does not raise this capacity. \rev{\citet{saunshi2025reasoning} find that looped models nearly match untied models of equal effective depth on synthetic reasoning tasks, gain more on reasoning than on memorization benchmarks, and can simulate $m$ chain-of-thought steps with $m$ loops. All of these results concern the output after the final loop.}

\paragraph{Adaptive depth.} \rev{Dynamic halting \citep{dehghani2019universal}, confidence-based early exit \citep{schuster2022confident}, token routing \citep{raposo2024mixture}, per-token recursion \citep{bae2025mixture} and learned exit gates \citep{zhu2025scaling} allocate depth per input. We propose no such mechanism. We measure, in a frozen model, when stored knowledge becomes readable, using a tuned lens \citep{belrose2023eliciting} and linear probes \citep{alain2018understandingintermediatelayersusing}.}

\paragraph{Latent multi-hop computation.} \citet{ltg} find a staircase between chain length and the recurrences needed for a correct endpoint. \citet{zhang2026knowinghopsenoughunderstanding} use patching to show that context-invariant bridge-entity representations support two-hop generalization, and \citet{fu2026discoloop} find that a bridge entity can be decodable yet misaligned with what the next recurrence expects. \rev{Concurrently, \citet{sesodia2026insideloop} probe weight-tied models on in-context graph traversal and find a staircase of intermediate nodes that patching can redirect; they leave open whether it persists when retrieval mixes with computation.} We study this case, with every hop retrieved from facts stored in the weights, and trace and patch successive bridge entities along longer chains.

\section{Limitations and Conclusions}
\paragraph{Limitations.}
Affine lens maps, linear probes and second-order geometry may miss nonlinear encodings, so unsuccessful readout does not establish that information is absent. Synthetic tasks provide controlled knowledge sources but do not establish whether the observed retrieval dynamics extend to natural-text models. We leave richer readouts, broader causal replication and extensions to natural text for future work.

\paragraph{Conclusions}
Final factual performance alone gives an incomplete account of how stored knowledge enters a computation. Our results distinguish when a fact becomes readable from whether its intermediate representation influences subsequent reasoning. In single-hop recall, comparable final scores conceal load-dependent retrieval delays that can be reduced through every-loop supervision with depth-specific readouts. In multi-hop composition, localized interventions connect intermediate readability to causal influence: replacing a probe-defined subspace of a bridge state redirects downstream entities and the final answer, while same-entity replacements largely preserve the answer across retrieval contexts. Together, these findings link a training-dependent retrieval process to the causal role of retrieved representations in composing stored facts. 

The broader implication is methodological: evaluating recurrent computation requires distinguishing final performance, intermediate readability, and causal influence. Depth-resolved readouts and controlled interventions address different parts of this distinction; neither final accuracy nor confidence alone resolves it. 
By tracking facts across recurrences and intervening on retrieved representations, our study makes the transition from stored knowledge to its use in reasoning directly testable. This moves the analysis of looped models beyond how much they remember to when and how that knowledge contributes to computation.

\subsection*{AI use statement}

We used a large language model as a writing and analysis aid: for prose editing of
sections drafted by the authors, and for exploratory analysis scripts over result
files that the authors then verified. Research questions, experimental design,
model training, and all reported measurements are the authors' own. Every number
in the paper was recomputed from the logged outputs by an author, and all
LLM-assisted code was read and tested before use. We take responsibility for the
final content of this work.

\subsection*{Ethics statement}

This work uses language models trained on synthetic data. No human subjects, personal data, or scraped text are involved, and the individuals studied do not exist. We see no direct pathway to harm from the measurements reported here. 

\subsection*{Reproducibility statement}

Sections~\ref{sec:setup} and \ref{sec:setup_multihop}\rev{ and Appendices~\ref{app:training} and~\ref{app:mh_seeds}} specify the model architectures, recurrence budgets, datasets, and training objectives. We fully describe the construction, fitting, and evaluation of the linear lenses and intermediate-entity probes, including all data splits and masking procedures. Equations~\ref{eq:fratio} and \ref{eq:eta} define the variance-ratio analysis and its permutation normalization, while Section~\ref{sec:causal_interventions} details the intervention protocol. Code and checkpoints will be released.

\bibliography{iclr2027_conference}

@misc{zhang2026knowinghopsenoughunderstanding,
      title={Why Knowing Both Hops Is Not Enough: Understanding Two-Hop Generalization in Language Models}, 
      author={Zili Zhang and Yilin Wang and Heng Wang and Herun Wan and Minnan Luo},
      year={2026},
      eprint={2608.07261},
      archivePrefix={arXiv},
      primaryClass={cs.CL},
      url={https://arxiv.org/abs/2608.07261}, 
}

@misc{alain2018understandingintermediatelayersusing,
      title={Understanding intermediate layers using linear classifier probes}, 
      author={Guillaume Alain and Yoshua Bengio},
      year={2018},
      eprint={1610.01644},
      archivePrefix={arXiv},
      primaryClass={stat.ML},
      url={https://arxiv.org/abs/1610.01644}, 
}

@inproceedings{saunshi2025reasoning,
  title={Reasoning with latent thoughts: On the power of looped transformers},
  author={Saunshi, Nikunj and Dikkala, Nishanth and Li, Zhiyuan and Kumar, Sanjiv and J Reddi, Sashank},
  booktitle={International Conference on Learning Representations},
  volume={2025},
  pages={14855--14881},
  year={2025}
}

@misc{xu2020theoryusableinformationcomputational,
  title={A theory of usable information under computational constraints},
  author={Xu, Yilun and Zhao, Shengjia and Song, Jiaming and Stewart, Russell and Ermon, Stefano},
  journal={arXiv preprint arXiv:2002.10689},
  year={2020}
}

@inproceedings{dehghani2019universal,
  title={Universal Transformers},
  author={Dehghani, Mostafa and Gouws, Stephan and Vinyals, Oriol and Uszkoreit, Jakob and Kaiser, {\L}ukasz},
  booktitle={International Conference on Learning Representations},
  year={2019}
}

@article{schuster2022confident,
  title={Confident adaptive language modeling},
  author={Schuster, Tal and Fisch, Adam and Gupta, Jai and Dehghani, Mostafa and Bahri, Dara and Tran, Vinh and Tay, Yi and Metzler, Donald},
  journal={Advances in Neural Information Processing Systems},
  volume={35},
  pages={17456--17472},
  year={2022}
}

@article{raposo2024mixture,
  title={Mixture-of-depths: Dynamically allocating compute in transformer-based language models},
  author={Raposo, David and Ritter, Sam and Richards, Blake and Lillicrap, Timothy and Humphreys, Peter Conway and Santoro, Adam},
  journal={arXiv preprint arXiv:2404.02258},
  year={2024}
}

@article{belrose2023eliciting,
  title={Eliciting latent predictions from transformers with the tuned lens},
  author={Belrose, Nora and Ostrovsky, Igor and McKinney, Lev and Furman, Zach and Smith, Logan and Halawi, Danny and Biderman, Stella and Steinhardt, Jacob},
  journal={arXiv preprint arXiv:2303.08112},
  year={2023}
}

@article{bae2025mixture,
  title={Mixture-of-recursions: Learning dynamic recursive depths for adaptive token-level computation},
  author={Bae, Sangmin and Kim, Yujin and Bayat, Reza and Kim, Sungnyun and Ha, Jiyoun and Schuster, Tal and Fisch, Adam and Harutyunyan, Hrayr and Ji, Ziwei and Courville, Aaron and others},
  journal={Advances in Neural Information Processing Systems},
  volume={38},
  pages={96572--96617},
  year={2026}
}

@article{zhu2025scaling,
  title={Scaling Latent Reasoning via Looped Language Models},
  author={Zhu, Rui-Jie and Wang, Zixuan and Hua, Kai and Zhang, Tianyu and Li, Ziniu and Que, Haoran and Wei, Boyi and Wen, Zixin and Yin, Fan and Xing, He and others},
  journal={arXiv preprint arXiv:2510.25741},
  year={2025}
}

@article{allen2023physics,
  title={Physics of language models: Part 3.1, knowledge storage and extraction},
  author={Allen-Zhu, Zeyuan and Li, Yuanzhi},
  journal={arXiv preprint arXiv:2309.14316},
  year={2023}
}

@article{barber2004algorithm,
  title={The im algorithm: a variational approach to information maximization},
  author={Barber, David and Agakov, Felix},
  journal={Advances in neural information processing systems},
  volume={16},
  number={320},
  pages={201},
  year={2004}
}

@article{sesodia2026insideloop,
  title={Inside the Loop: A Mechanistic Study of Weight-Tied Transformers on Depth-Bound Algorithmic Tasks},
  author={Sesodia, Magnus and Sock, Juil and Torr, Philip and de Witt, Christian Schroeder},
  journal={arXiv preprint arXiv:2610.24135},
  year={2026}
}

@article{ltg,
  title={Loop, think, \& generalize: Implicit reasoning in recurrent-depth transformers},
  author={Kohli, Harsh and Parthasarathy, Srinivasan and Sun, Huan and Yao, Yuekun},
  journal={arXiv preprint arXiv:2604.07822},
  year={2026}
}

@article{fu2026discoloop,
  title={DiscoLoop: Looping Discrete Embeddings and Continuous Hidden States for Multi-hop Reasoning},
  author={Fu, Hengyu and Guo, Tianyu and Wang, Zixuan and Zhu, Hanlin and Lee, Jason D and Jiao, Jiantao and Russell, Stuart and Mei, Song},
  journal={arXiv preprint arXiv:2607.00341},
  year={2026}
}

@inproceedings{geva2023dissecting,
  title={Dissecting recall of factual associations in auto-regressive language models},
  author={Geva, Mor and Bastings, Jasmijn and Filippova, Katja and Globerson, Amir},
  booktitle={Proceedings of the 2023 Conference on Empirical Methods in Natural Language Processing},
  pages={12216--12235},
  year={2023}
}

@inproceedings{biran2024hopping,
  title={Hopping too late: Exploring the limitations of large language models on multi-hop queries},
  author={Biran, Eden and Gottesman, Daniela and Yang, Sohee and Geva, Mor and Globerson, Amir},
  booktitle={Proceedings of the 2024 Conference on Empirical Methods in Natural Language Processing},
  pages={14113--14130},
  year={2024}
}
\bibliographystyle{iclr2027_conference}

\appendix
\section{Storage capacity of our checkpoints}\label{app:capacity}

\paragraph{\rev{Model and training details.}}\label{app:training}\rev{ The shared block has 11 layers with 3 attention heads, $d=192$, rotary position embeddings, and a vocabulary of 5{,}086--5{,}371 tokens (it depends on $N$ and seed). We train with AdamW ($\beta_1=0.9$, $\beta_2=0.98$, weight decay $0.02$), a peak learning rate of $10^{-3}$ with 1{,}000 linear warmup steps and cosine decay to $10\%$ of the peak, an effective batch of 384 packed sequences of 512 tokens (split into micro-batches by GPU memory: $64\times3$ for loop-8 and $96\times2$ for loop-4 in seed 0, $96\times2$ and $192\times1$ in seeds 1--4), bf16 precision, and 350 exposures per biography, i.e.\ 17{,}318, 43{,}295 and 86{,}589 optimizer steps at $N=$ 20K, 50K and 100K for both loop budgets.}

\begin{figure}[!ht]
\centering
\includegraphics[width=0.5\linewidth]{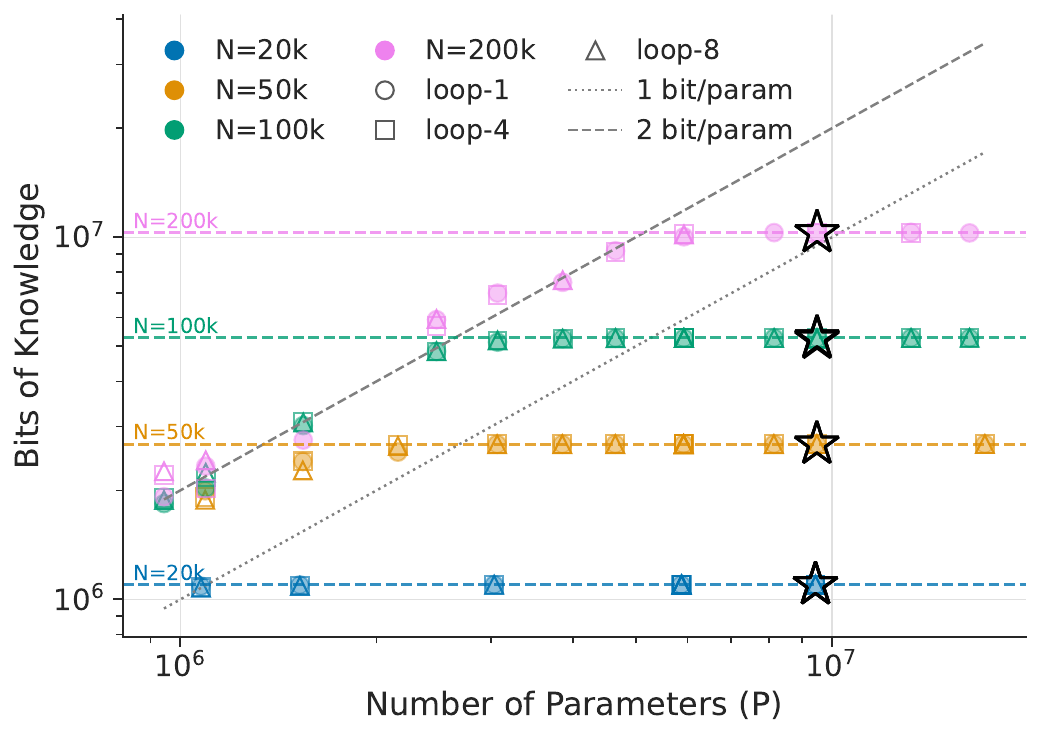}
\caption{\textbf{Looping does not change what is stored.} Total memorised bits
against parameter count for models trained on BioS, colours by dataset size and
marker shapes by loop budget. Dashed horizontal lines are the ceiling $R_{\max}$
for each dataset. Loop-1, loop-4 and loop-8 coincide everywhere. Stars mark the
models used throughout the paper: identical architecture across dataset sizes, all
at their ceiling.}
\label{fig:capacity}
\end{figure}

\paragraph{Storage capacity.} 
Figure~\ref{fig:capacity} reproduces Figure 6 of \cite{zhu2025scaling} and shows highly similar results.

\newpage

\section{Lens capacity sweep}\label{app:ranks}

\begin{figure}[!ht]
\centering
\includegraphics[width=\linewidth]{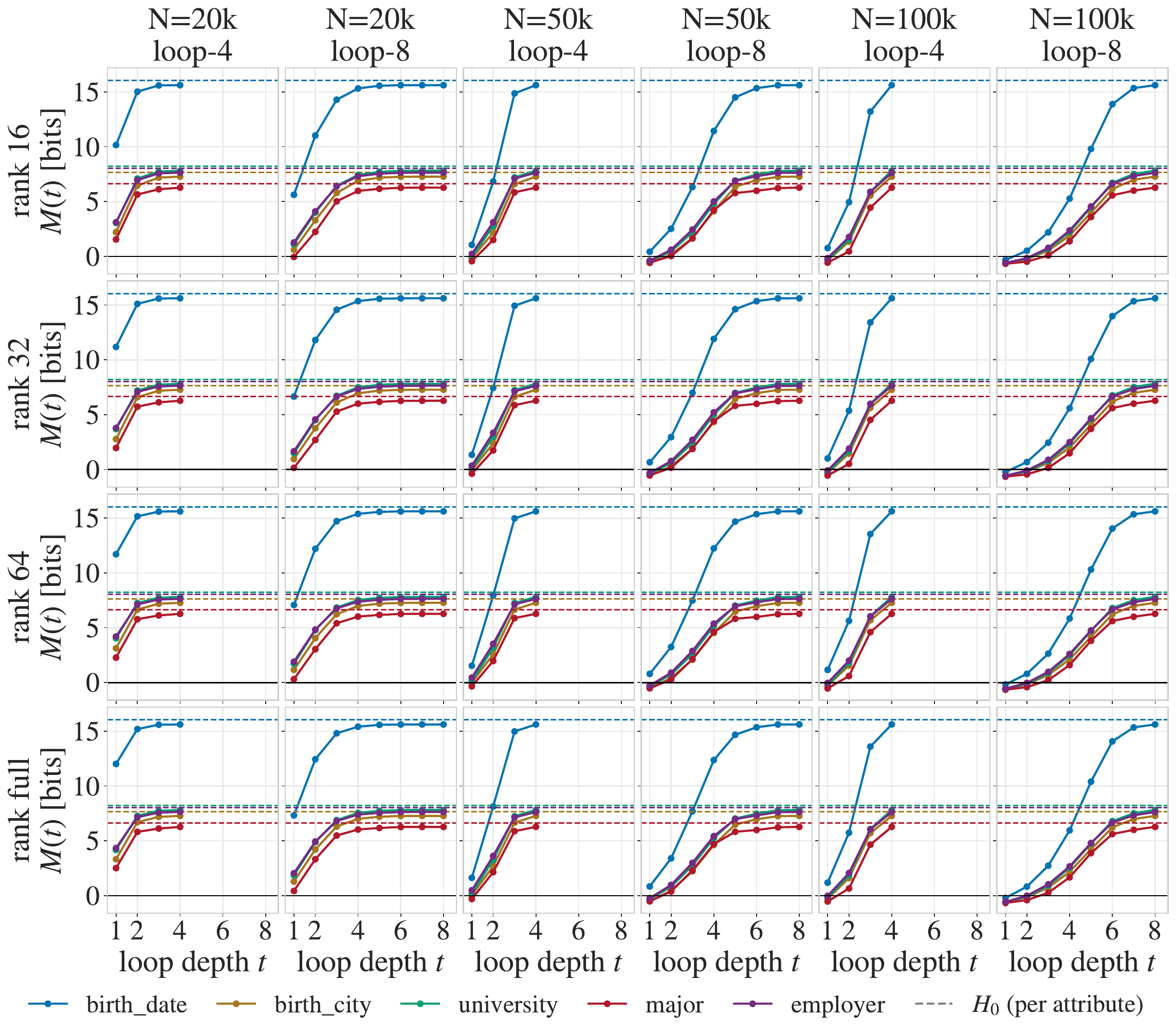}
\caption{
\textbf{Lens capacity and readable bits.} $M(t)$ per attribute for six $(N,L)$ cells, seed 0. Rows: update ranks 16, 32 and 64 in $W_t=I+U_tV_t$, then an unrestricted affine lens map. Dashed: source entropy $H_0$; black: $M=0$. All rows use learning rate $10^{-3}$.}
\label{fig:ranks}
\end{figure}

Figure~\ref{fig:ranks} varies the capacity of the affine lens map in Section~\ref{sec:setup}. The restricted variants use $W_t=I+U_tV_t$, where $U_t\in\mathbb R^{192\times r}$ and $V_t\in\mathbb R^{r\times192}$ for $r\in\{16,32,64\}$. The update has rank at most $r$; $W_t$ is generally full rank, and the identity path remains available. Including bias, these variants have $6{,}336$, $12{,}480$ and $24{,}768$ fitted parameters per depth, versus $37{,}056$ for the unrestricted affine map and none for the native readout. The qualitative load-dependent schedule persists, though early values and first-crossing depths change. The rank sweep uses seed 0 and a fixed learning rate of $10^{-3}$ for every row, including the unrestricted row; the main full-lens results instead select learning rates per depth (Appendix~\ref{app:lens_controls}).

\textbf{How to read the figure.} Each column is one $(N,L)$ cell. Rows show update ranks 16, 32 and 64, then the unrestricted affine map. Coloured curves show attributes; matching dashed lines mark source entropy $H_0$, reached at zero evaluated CE. The solid black line marks the uniform-value log-loss baseline, $M=0$.

\textbf{Update rank mainly changes early readability.} Final values are nearly identical across the four variants. In seed 0, the unrestricted lens exceeds the rank-16 update by at most $2.1$ bits for any attribute and depth. The first sampled loop reaching $95\%$ of the five-attribute mean's own final value is unchanged except at (20K, loop-4), where update ranks 16 and 32 cross one loop later. The interpolated crossings there are $2.28$, $2.10$, $2.00$ and $1.99$ loops. For birth date at $t=1$, $M(1)$ increases across the four variants: approximately $10.0/11.1/11.8/12.0$ bits at (20K, loop-4), $4.3/5.1/5.5/5.7$ at (20K, loop-8), and $4.0/4.7/5.2/5.4$ at (50K, loop-4).

\textbf{Load dependence persists across update ranks.} At 20K with loop-4, the curves rise early; larger loads delay the rise. At 100K, in seed 0, the single-token attributes are below $M=0$ at $t=1$ in both budgets and at $t=2$ with loop-8, across all variants ($-1.05$ to $-0.36$ bits). With loop-4 they become positive at $t=2$ ($0.20$--$1.16$ bits); with loop-8 they are near zero at $t=3$ ($-0.25$ to $0.41$ bits) and positive from $t=4$ ($1.02$--$1.80$ bits). Moving from loop-4 to loop-8 at fixed $N$ spreads the rise over more depths without a material change in final bits.

\textbf{Comparing attributes.} The four single-token attributes follow closely spaced curves, with different final levels reflecting their different source entropies. Birth date has a larger $H_0$, so its larger absolute score should not by itself be interpreted as earlier retrieval. In the unrestricted-lens loop-8 cells, its rise also occurs earlier relative to its own $H_0$.

\begin{figure}[!ht]
\centering
\includegraphics[width=\linewidth]{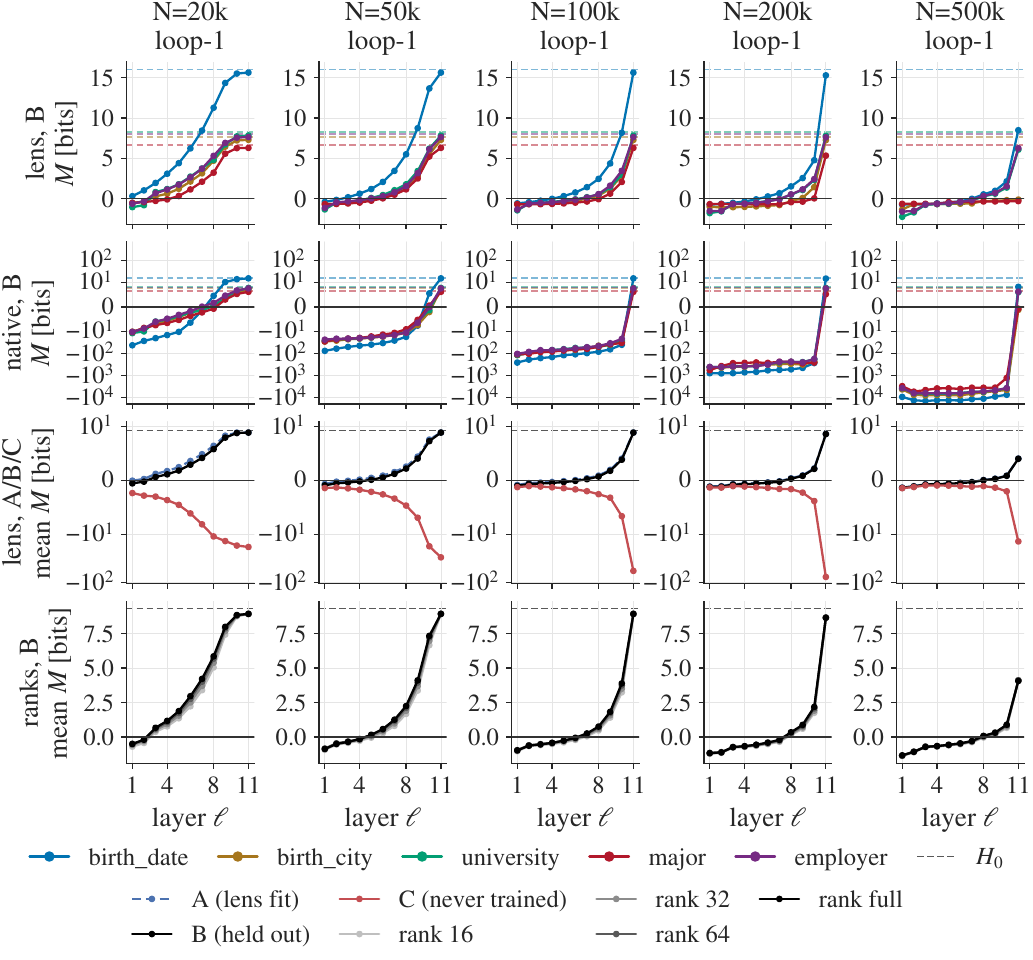}
\caption{
\textbf{Per-layer readout of loop-1 models} (11 layers, $d=192$; one run per $N$, seed 0). Rows: full lens on B per attribute ($H_0$ dashed); native readout on B; lens on A, B and C; lens on B with update ranks 16, 32, 64 and an unrestricted lens map. Rows 2--3 use a symmetric-log axis, linear within $\pm10$ bits. Bands: 95\% bootstrap intervals over individuals, not training seeds.}
\label{fig:perlayer_loop1}
\end{figure}

\textbf{Single-block models, read per layer.} Figure~\ref{fig:perlayer_loop1} fits the lens per layer for the loop-1 models with the 11-layer, $d=192$ block (one run per $N$, seed 0). The five-attribute mean first becomes positive at layer 3, 5, 7, 8 and 8 for 20K, 50K, 100K, 200K and 500K. At 500K, final $M$ is $4.06$ bits against mean source entropy $9.32$ bits. The native readout becomes positive later than the lens at every $N$; from 100K onwards it is strongly negative before the last layer. On C the lens remains below the uniform-value log-loss baseline at every layer. Low-rank updates to identity with ranks 16, 32 or 64 change the interpolated first depth reaching $95\%$ of the readout's own final value by at most $0.23$ layers relative to the unrestricted lens.

\FloatBarrier
\section{Lens fitting protocol and decoder controls}\label{app:lens_controls}

\textbf{Splits.} Individuals are split into three disjoint groups: A ($60\%$), B ($35\%$), and C ($5\%$), i.e.\ $12{,}000/7{,}000/1{,}000$ individuals at 20K, $30{,}000/17{,}500/2{,}500$ at 50K, and $60{,}000/35{,}000/5{,}000$ at 100K. No name appears in more than one group. The split is by person, not by paraphrase. Each person's attributes are drawn independently and uniformly, as in BioS \citep{allen2023physics}, so the names in A carry no information about the attribute values in B.

\textbf{Fitting and scoring.} At each depth, $(W_t,b_t)$ starts at identity and zero bias and minimizes gold attribute-token CE; the backbone, terminal LayerNorm and embedding-tied head are frozen. This differs from the final-output KL distillation of \citet{belrose2023eliciting}. Each individual has one biography with randomly chosen templates and sentence order. Readouts predict attribute-value tokens under teacher forcing, using a full-vocabulary softmax and unclipped CE in bits. Earlier attributes may already occur in the prefix; there is no separate post-loop query. Birth-date CE sums three token losses with gold prefixes. We fit on $95\%$ of A; the other $5\%$ selects each depth's learning rate from $\{3\cdot10^{-4},10^{-3},3\cdot10^{-3},10^{-2}\}$ and best epoch (AdamW, 30 epochs, batches of 1{,}024 cached scored-token rows, $10\%$ linear warmup followed by cosine decay). Nothing is fit or selected on B or C. Reported A curves include both fit and validation individuals; the main readability results use B.

\textbf{Score interpretation.} $M=H_0-\mathrm{CE}$ uses the entropy of the uniform generating source, not the empirical entropy of the fixed evaluation population. The Barber--Agakov inequality \citep{barber2004algorithm} is $I(X;H)\geq H(X)+\mathbb E\log_2 q(X\mid H)$, with entropy and expectation under the same specified distribution. Replacing $H(X)$ by source entropy need not give an MI lower bound on a fixed population: at 20K, B has 7,000 individuals, so its empirical birth-date entropy cannot exceed $\log_2 7000\approx12.77$ bits, below the source entropy $\log_2 67200\approx16.04$. Moreover, date CE uses three predictive states. We therefore interpret $M$ operationally as predictive log-loss reduction relative to the source baseline, without identifying it unconditionally with MI or compression-based capacity.

\textbf{Controls.} Table~\ref{tab:lens_controls} reports the full lens on the 30 main models (11-layer block, $d=192$; five seeds per cell). For the five-attribute mean, A exceeds B by less than $0.71$ bits at every depth and at most $0.017$ bits at the final depth. Mean interpolated first crossings of $95\%$ of each curve's own final value on A are $1.92/2.92/3.75$ loops (loop-4) and $3.53/5.28/6.65$ (loop-8) at 20K/50K/100K; on B they are $1.99/2.98/3.77$ and $3.86/5.53/6.73$ (five-seed means). The load ordering is unchanged; no individual model shifts by more than $0.62$ loops. C remains below the uniform-value log-loss baseline for every attribute and depth: its largest score is $-0.80$ bits for an attribute and $-1.10$ for the five-attribute mean. These controls support generalization beyond lens-fitting individuals and distinguish backbone-trained from never-trained populations; they do not prove that fitting the decoder contributes no information. Figure~\ref{fig:lens_controls} shows both controls across depth.

\begin{table}[!ht]
\centering
\caption{
Decoder controls for the full lens (five-attribute mean $M(t)$ in bits; Figure~\ref{fig:lens_controls}). A includes fit and validation individuals; B is held out from lens fitting but seen by the backbone. A$-$B: largest gap over depths and seeds; C: largest score over depths and seeds. B: five-seed mean ($\pm$SD at $t=1$; final SD at most $0.005$).}
\label{tab:lens_controls}
\small
\begin{tabular}{lcccc}
\toprule
Cell & max$_t$ (A$-$B) & B at $t=1$ & B at $t=L$ & max$_t$ C \\
\midrule
20K, loop-4 & 0.69 & $5.67\pm0.76$ & 8.93 & $-9.57$ \\
20K, loop-8 & 0.70 & $2.32\pm0.38$ & 8.93 & $-4.76$ \\
50K, loop-4 & 0.39 & $1.21\pm0.53$ & 8.93 & $-2.34$ \\
50K, loop-8 & 0.38 & $-0.09\pm0.11$ & 8.93 & $-1.47$ \\
100K, loop-4 & 0.21 & $-0.11\pm0.11$ & 8.93 & $-1.41$ \\
100K, loop-8 &  0.21 & $-0.56\pm0.07$ & 8.93 & $-1.10$ \\
\bottomrule
\end{tabular}
\end{table}

\FloatBarrier

\subsection{Supervision at every loop and an exit rule}\label{app:every_loop}

\begin{figure}[!ht]
\centering
\includegraphics[width=\linewidth]{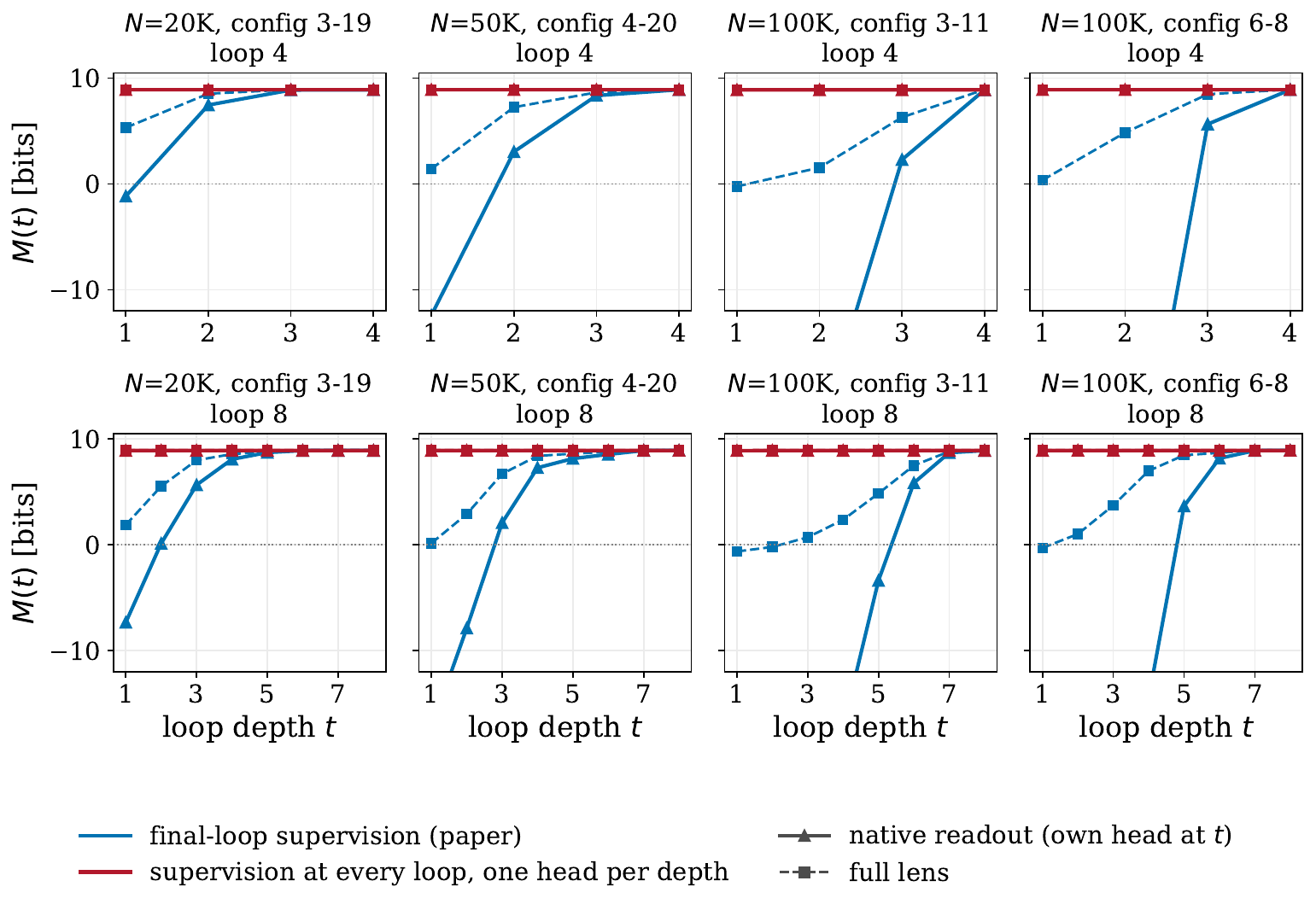}
\caption{
\textbf{Every-loop supervision with depth-specific readouts.} $M(t)$ on B for final-loop models (blue) and every-loop twins (red); one run per arm per cell. Solid: own readout; dashed: full lens. The axis is clipped at $-12$ bits; final-loop native curves reach $-94$ at $t=1$.}
\label{fig:every_loop_supervision}
\end{figure}

\textbf{Every-loop twins.} For eight cells, we compare final-loop training with a twin using the unweighted mean of CE over all $L$ depths. Each depth has its own LayerNorm and decodes through the tied embedding matrix, with no new $d\times V$ output weights. Thus supervision and readouts both change; all other training settings are matched (the saved configurations differ only in these flags and output directory). Cells are a 19-layer, $d=192$ block at 20K, a 20-layer, $d=256$ block at 50K, and at 100K the main 11-layer, $d=192$ block and an 8-layer, $d=384$ block, each at loop-4 and loop-8, with one run per arm per cell. We measure five-attribute $M(t)$ on B using each model's own readout and a full lens, plus the first depth reaching $95\%$ of the readout's own final value. The value-score midpoint is available for both arms at 100K with the 11-layer block. Figure~\ref{fig:every_loop_supervision} compares the protocols.

\textbf{Every-loop results.} In all eight cells, the twin's own readout gives $M(1)=8.89$--$8.90$ bits, within $0.02$ bits of its final value, and the lens gives $8.92$--$8.93$ bits; both first cross $95\%$ of their own final value at loop 1. Final-loop models start at $-94.1$ to $-1.2$ bits (own readout) and $-0.6$ to $5.3$ (lens), first reaching $95\%$ of their final lens score after $2.0$--$6.8$ loops. Final $M$ differs between arms by less than $0.011$ bits. At 100K with the 11-layer block, the normalized geometric midpoint moves from $0.72$ to $0.25$ for loop-4 and from $0.64$ to $0.13$ for loop-8, the first sampled loop in each budget. These first-sample values are upper bounds on the midpoint depth. Near-final first-loop readability is therefore achievable under this combined supervision-and-readout protocol; delayed readability is not unavoidable in these settings.

\begin{figure}[t]
\centering
\includegraphics[width=\linewidth]{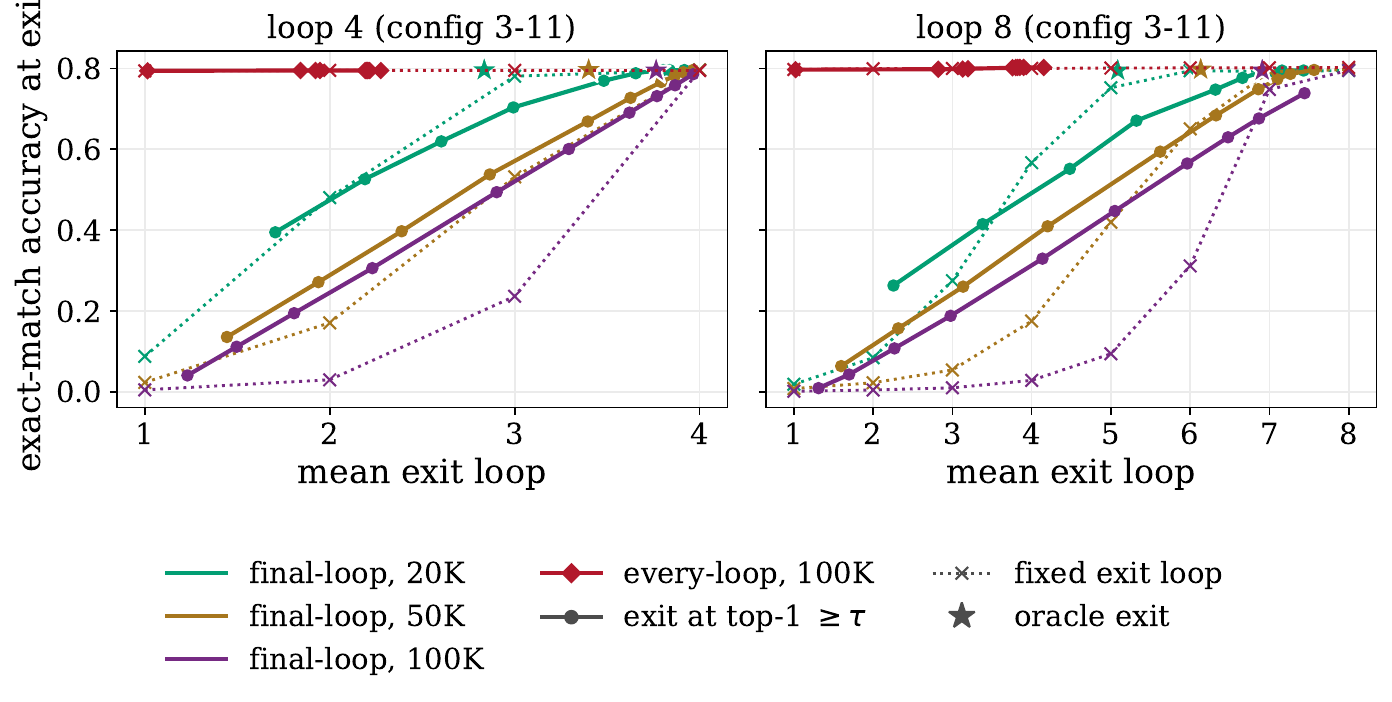}
\caption{
\textbf{Confidence-threshold exit under teacher forcing.} Accuracy versus mean exit loop on B; one run per cell. Circles: final-loop models at 20K, 50K and 100K; red diamonds: 100K every-loop twins; dotted: fixed exit; stars: oracle. Full-depth accuracy is about $0.80$. Threshold-rule queries that never cross $\tau$ use $L$ loops; separately, the oracle charges $L$ for queries wrong at the final loop.}
\label{fig:exit_rule}
\end{figure}

\textbf{Exit rule.} Figure~\ref{fig:exit_rule} tests a confidence-threshold rule: exit at the first loop where the native maximum token probability, averaged over answer tokens, is at least $\tau$, otherwise run all $L$ loops. We test a threshold grid from $0.5$ to $0.999$ on individual--attribute pairs in B (35K--175K queries per model), using the main 11-layer block at 20K, 50K and 100K, one run per cell (seed 0). Evaluation is teacher-forced. In particular, birth-date confidence and joint token accuracy use gold intermediate tokens; these measurements do not evaluate a free-generation date-exit policy. The four single-token attributes do not require such intermediate tokens. The oracle exits at the earliest loop from which all later answers remain correct; queries wrong at $L$ are charged $L$ loops. We do not test lens-based exit.

\textbf{Exit-rule results.} Full-depth exact-match accuracy is $0.795$--$0.803$, including twins; near-ceiling readable bits do not imply near-perfect accuracy. For final-loop models at 20K and 50K, the smallest tested threshold within $0.5$ percentage points of full-depth accuracy uses $3.76$--$3.97$ of 4 loops and $6.90$--$7.56$ of 8, versus oracle means $2.84$--$3.40$ and $5.09$--$6.13$. At 100K no tested threshold qualifies (best: $0.786$ versus $0.795$ for loop-4; $0.739$ versus $0.795$ for loop-8). At $\tau=0.5$, these models exit after $1.2$--$1.3$ loops with accuracy $0.04$ and $0.01$: early native confidence is poorly aligned with correctness. This conclusion concerns the tested grid, not all possible thresholds. For the 100K every-loop twins, fixed first-loop exit loses approximately $0.078$ and $0.502$ percentage points for loop-4 and loop-8. At $\tau=0.99$, mean exit depths are $2.21$ and $3.85$; among queries wrong at full depth, 0 of 35{,}874 (loop-4) and 7 of 34{,}533 (loop-8) ever reach this threshold. Correct-at-final queries exit after $1.75$ and $2.83$ loops on average.

\section{Further details of the multi-hop experiment} \label{app:fur_mh}

This appendix provides complementary analyses and additional details for the multi-hop experiments presented in Section~\ref{sec:multihop}.

\subsection{Information retrieval in multi-hop chains}

\paragraph{Task and probe details.} In the prompt of Figure~\ref{fig:mh_prompt}, the start entity is followed by one relation token per hop. The text target appends the endpoint and \texttt{</a>}, but the data loader removes \texttt{</a>} and the training loss scores only the endpoint token; intermediate entities are not separately supervised outputs. The bridge entities are obtained by walking the atomic facts of the training set. The same four-layer block is applied $R$ times ($R=6$ for R6, sampled for R\_dyn), and the answer is read from the final state through the final LayerNorm and the tied output head. To measure intermediate-entity readability, we run each checkpoint for 10 loops and extract states after every physical layer, applying the model's final LayerNorm before fitting affine probes over the 217-token vocabulary. The probe data come from the test split of \citet{ltg}, with chain lengths 2--25. Targets are intermediate entities at hop indices 1--15, excluding each chain's endpoint. Chains are split 80/10/10 into probe-training, validation and test sets, stratified by chain length and source type. The dynamic-model feature cache contains 52{,}800/6{,}600/6{,}600 chains in these splits. Inputs are padded to 50 token positions, with an attention mask identifying real tokens. Each probe reads one token position rather than concatenating positions; separate probes are fit for each loop, layer, hop and token position, for 10 epochs.

\begin{figure}[t]
  \centering
  \includegraphics[width=\linewidth]{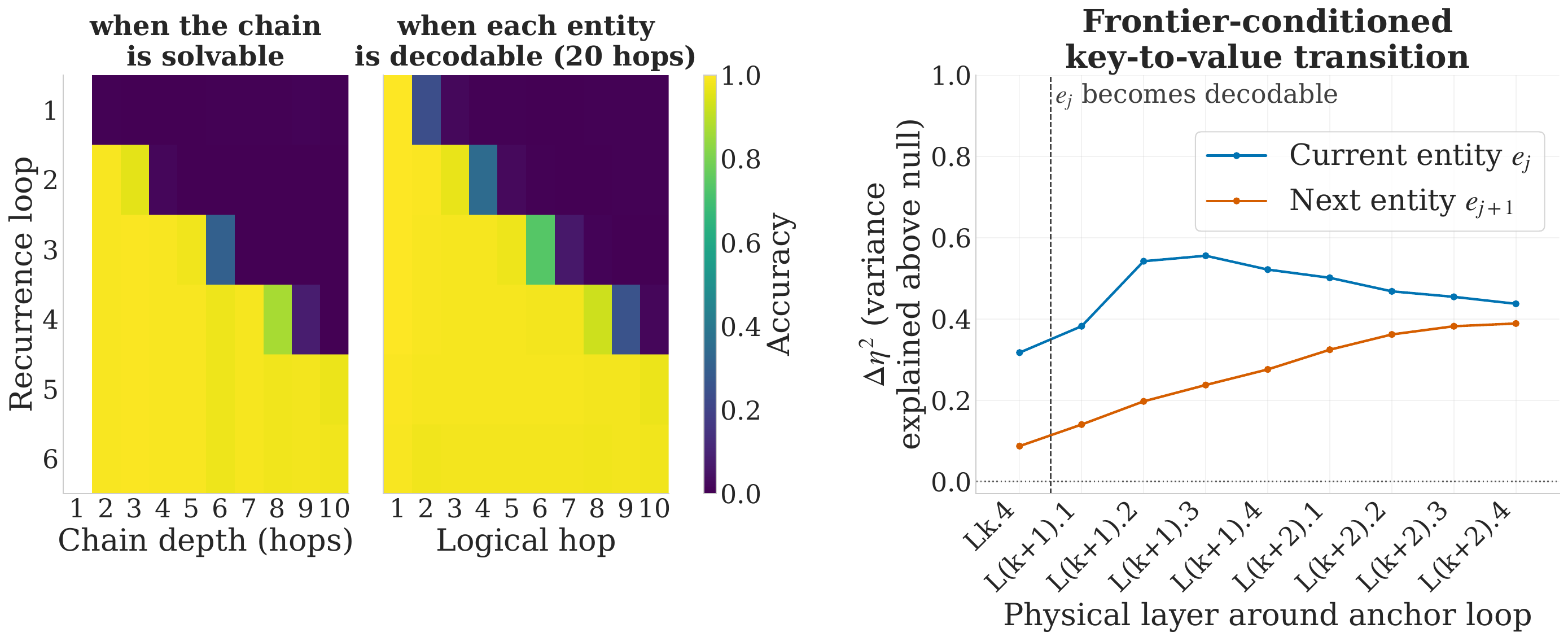}
  \caption{\textbf{(Left)} Released fixed-loop checkpoint: endpoint accuracy by chain length and intermediate-entity decodability on 20-hop chains, for loops 1--6 and hops 1--10. The two heatmaps show a similar staircase. \textbf{(Right)} Difference between entity-grouped $\eta^2$ and its permutation-null value, $\Delta\eta^2$, for 1{,}069 selected chains. The window begins at layer 4 of loop 2, where $e_4$ is newly decoded. Current-entity structure rises and then declines, while next-entity structure grows without overtaking it. This is a difference of variance-explained scores, not itself a variance fraction.}
  \label{fig:acc_vs_entity_and_kv_multihop}
\end{figure}

\paragraph{Geometry of a selected bridge transition.} In a reasoning chain, a bridge entity is the value retrieved at one hop and, together with the next relation, the key for the following hop. As a complement to the ordered-decoding analysis in Section~\ref{sec:multi_hop_retrieval}, we extend the key--value geometry analysis of Section~\ref{sec:keyvalue} to ask whether next-entity structure develops while current-entity structure remains present. We follow the transition from $e_4$ to $e_5$ in 1{,}069 chains from the released fixed-loop checkpoint (R6) of \citet{ltg}. These are chains where $e_4$ is the furthest bridge decoded at layer 4 of loop 2 but not at layer 3. At the anchor and each subsequent layer, we concatenate the states at the positions of $r_4$ and $r_5$, subtract the mean within each next-relation group, and compute the null-corrected scores $\Delta\eta^2$ for the current and next entities using permutations within those groups. The current-entity score rises from 0.05 at the anchor to 0.63 at layer 3 of loop 3, then falls to 0.42 by layer 4 of loop 5; the next-entity score rises from 0.003 to 0.40 over the same interval (Figure~\ref{fig:acc_vs_entity_and_kv_multihop}, right). Thus, next-entity structure develops while current-entity structure persists, paralleling the overlap between key and value structure in single-hop retrieval. This selected-cohort analysis does not establish the same transition at every hop or demonstrate that the current representation drives the next retrieval; causal influence is tested separately in Section~\ref{sec:causal_interventions}.

\textbf{Endpoint accuracy per loop and hop length.} To test how additional inference loops extend the solved range, we evaluate 750 test chains per hop length from \citet{ltg} over hops 2--25 and loops 1--10. This sweep is separate from the smaller heatmap in Figure~\ref{fig:acc_vs_entity_and_kv_multihop}; Figure~\ref{fig:mh_seed_first_loop} summarizes first crossings for hops 2--13. We first average accuracy over the released checkpoint and three retrained seeds per regime, then apply the 80\% threshold. For fixed-loop training, the longest solved chain reaches 15 hops at loop 7; loops 8--10 do not increase this maximum, and it falls to 14 at loops 9--10. For dynamic training, the longest solved chain increases at every loop, reaching 25 hops at loop 10. The dynamic pool combines the released 22-hop-stage checkpoint with retrained 19-hop-stage checkpoints. Individual checkpoints reach maxima of 14--17 hops (fixed) and 24--25 hops (dynamic) within 10 loops.

\begin{figure}[t]
  \centering
  \includegraphics[width=\linewidth]{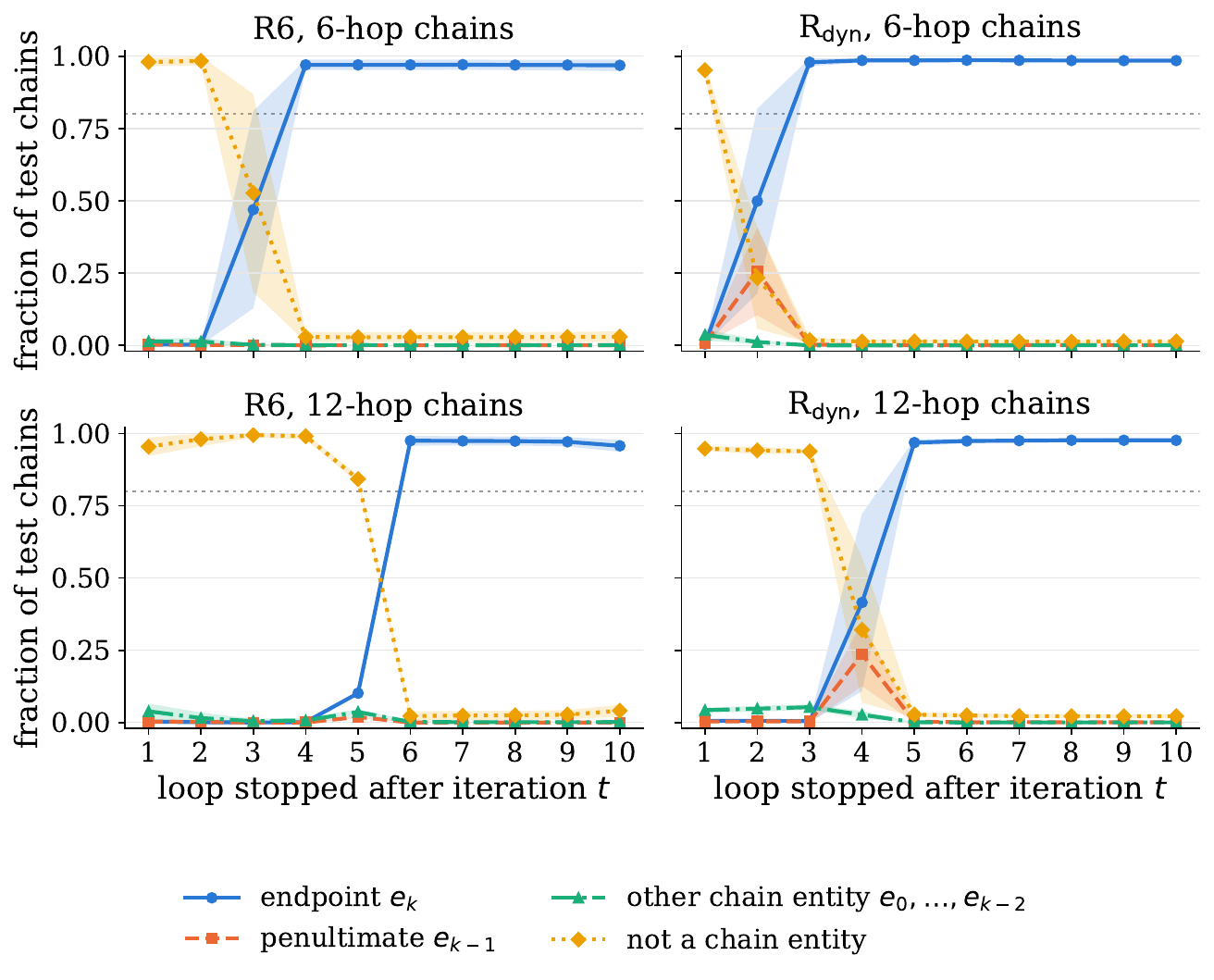}
  \caption{\textbf{What the model outputs after each loop.} Share of test chains whose own-head top-1 token after $t$ loops is $e_k$ (blue), $e_{k-1}$ (orange), $e_0,\ldots,e_{k-2}$ (green), or outside the chain (yellow). Top: 6 hops; bottom: 12 hops; left: fixed-loop training; right: dynamic training. Mean $\pm1$ SD over four checkpoints: one released and three retrained. Fixed checkpoints are at stage 13; dynamic checkpoints are at stage 22 (released) or 19 (retrained). Dotted line: 80\% reference.}
  \label{fig:mh_ownhead}
\end{figure}

\textbf{Own-head readout of the chain entities.} The probes of Figure~\ref{fig:multihop_r6_vs_rd} are trained decoders. As a check that uses no trained decoder, we stop the loop after $t$ iterations and read the model's own final LayerNorm and tied output head at the position scored by the evaluation of \citet{ltg}, on all 750 6-hop and all 750 12-hop test chains. We do this for the published R6 and R\_dyn checkpoints and for three retrained seeds of each regime (R6 seeds stopped at the 13-hop curriculum stage, R\_dyn seeds at the 19-hop stage). At every loop, its endpoint accuracy matches the corresponding saved endpoint-accuracy sweep. Figure~\ref{fig:mh_ownhead} shows the share of chains whose top-1 token is the endpoint $e_k$, the penultimate entity $e_{k-1}$, another chain entity ($e_0,\ldots,e_{k-2}$), or a token outside the chain. Some chains visit the same entity more than once, so each chain is counted once, in this order. In both regimes, the head writes the endpoint $e_k$ once the chain is solved. It rarely writes the start entity $e_0$ or the bridges $e_1,\ldots,e_{k-2}$: at most 6.7\% of chains for any checkpoint and loop, for this exclusive category. Entity repetitions mean that the number of distinct tokens in the category can vary across chains. The exception is the penultimate entity $e_{k-1}$ in R\_dyn. At loop 2 for 6-hop chains and loop 4 for 12-hop chains, the R\_dyn head writes $e_{k-1}$ in 6.0--42.5\% and 8.4--33.6\% of chains, depending on the checkpoint, whereas R6 does so in at most 2.7\% of chains at any loop. Thus, probe decodability does not generally coincide with an intermediate entity being the own-head top-1 prediction at this answer position. Dynamic training nevertheless produces a substantial penultimate-entity signal in some checkpoints.

\subsection{Training differences between the released checkpoints}\label{app:mh_seeds}

Table~\ref{tab:ltg_training} compares the two released checkpoints of \citet{ltg}. Both use a recurrent block of four layers with hidden size 768, 12 attention heads and no positional embeddings, a knowledge graph of 200 entities and 10 relations with 2k atomic facts and 15k $k$-hop facts per $k$, and AdamW with learning rate $10^{-4}$, weight decay 0.01, 2{,}000 linear warmup steps and batch 128 \citep[\S6.1, App.~C]{ltg}. The loss is the cross-entropy of the answer token after the last recurrence. Each curriculum stage creates a new optimizer and warmup schedule, and stage $k$ ends once held-out $k$-hop accuracy exceeds 95\%. The implementation uses linear warmup followed by linear decay within each stage, with the scheduler updated after every optimizer step. The curriculum gate checks the current hop length, rather than jointly rechecking every shorter length.

\begin{table}[t]
\centering
\caption{\textbf{Training differences between the released checkpoints.} Sources: [P] \citet{ltg}; [R] their training code; [E] our inspection or evaluation of the checkpoints.}
\label{tab:ltg_training}
\footnotesize
\setlength{\tabcolsep}{4pt}
\resizebox{\linewidth}{!}{\begin{tabular}{lccl}
\toprule
 & R6 & R\_dyn & Source \\
\midrule
Recurrences in training & always 6 & per batch, clip(Poisson(4), 2, 8) & [P, \S4, App.~J], [R] \\
Depths at which the answer is supervised & 6 & 2--8 & [P, \S4], [R] \\
Curriculum depth reached (hops) & 13 & 22 & [P, Fig.~5], [E] \\
Training facts in the last stage & 182{,}000 & 317{,}000 & [P, \S6.1], [E] \\
Optimizer steps per epoch, last stage & 1{,}422 & 2{,}477 & [P, App.~C], [E] \\
Cumulative epochs & 1{,}015 & 5{,}388 & [E] \\
Epochs / optimizer steps, last stage & 2 / 2{,}844 & 121 / 299{,}717 & [E] \\
Total optimizer steps & not available & $>1.3$M to learn up to 4 hops & [P, \S6.2] \\
Held-out accuracy $\geq95\%$ & 2--13 hops (6 loops) & 2--22 hops (8 loops) & [E] \\
\bottomrule
\end{tabular}}
\end{table}

\paragraph{Retrained seeds on matched data.} We trained three further seeds of each regime with the code, data and curriculum of \citet{ltg}; a seed changes the initialisation and the data order. To match the training data, we stop the R6 seeds and evaluate the R\_dyn seeds at the end of their 13-hop curriculum stage, where the released R6 stops; each such checkpoint is the last epoch before the curriculum moved to 14 hops. The R6 seeds stop after 788, 1{,}400 and 2{,}051 epochs (released model: 1{,}015); the R\_dyn seeds finish the 13-hop stage after 3{,}168--4{,}237 epochs and the 19-hop stage after 3{,}727--4{,}402 (released model: 22-hop stage, 5{,}388 epochs). On matched data (Figure~\ref{fig:mh_seed_first_loop}), R\_dyn solves 2- to 6-hop chains at the same loop as R6 or up to one loop earlier, 7- to 10-hop chains at the same loop (one seed is one loop later at 8 hops), 11- and 13-hop chains one loop later, and 12-hop chains at the same loop or one loop later. Trained further, to the 19-hop stage, the same R\_dyn seeds solve every hop length from 2 to 13 at the same loop as R6 or one loop earlier (e.g.\ 12 hops at loop 5 against 6). The longest hop length reaching 0.8 within 10 loops is 14--17 (R6), 16--17 (R\_dyn, 13-hop stage), 24--25 (19-hop stage) and 25 hops (released R\_dyn).

\paragraph{Solved range at the gated loop count.} To distinguish learned chain length from extrapolation beyond it, we also measure the solved range at the inference budget used for curriculum evaluation. In every checkpoint, the longest chain reaching 0.8 at the loop count on which the curriculum is gated (6 for R6, 8 for R\_dyn) exceeds the longest trained chain by at most three hops: 13 hops for all four R6 checkpoints, 14--15 for the R\_dyn seeds at the 13-hop stage, 21--22 at the 19-hop stage and 23 for the released R\_dyn (22-hop stage). Up to that loop, each further loop extends the solved range by 2--3 hops in R6, but by 1--3 hops (1.7--1.9 on average) in R\_dyn at the 13-hop stage, which therefore spreads the same 13-hop range over seven loops instead of six. At the 19-hop stage, the same R\_dyn seeds extend it by 2--3 hops per loop. The 0.95 gate enforces accuracy on the current curriculum hop length at this budget, not simultaneous accuracy on the entire trained range. The additional solved hops and the per-loop rates are measured outcomes, not consequences imposed by the gate.

\subsection{Causal patches on reasoning chains}
\label{app:further_counterfactual}

\begin{figure}[t]
  \centering
  \includegraphics[width=.72\linewidth]{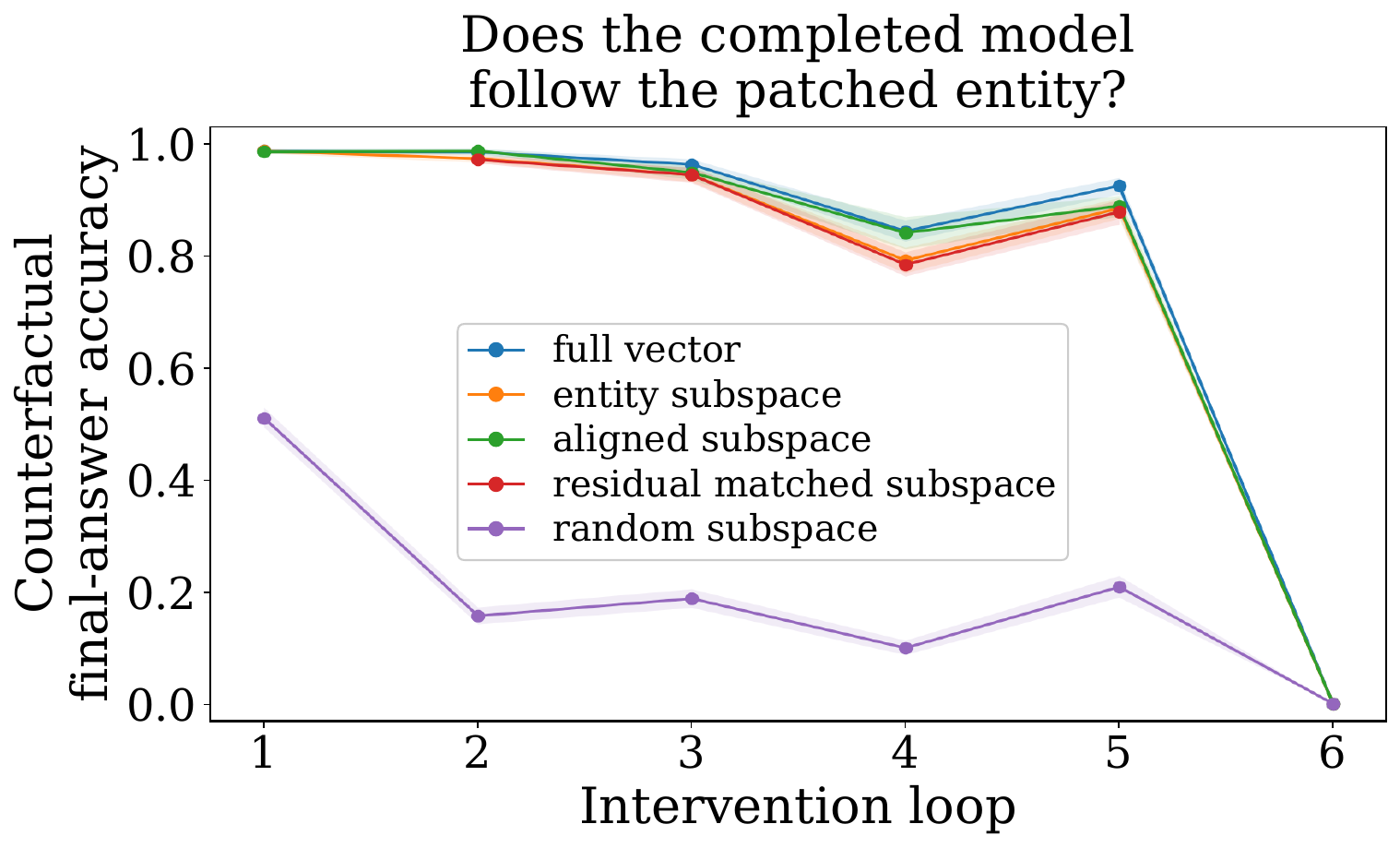}
  \caption{Final counterfactual-answer accuracy after patching at each
  recurrence and running the remaining loops.  The model's native LM head
  follows entity-subspace interventions, while random-subspace patches are less effective. One fixed-loop checkpoint; three evaluation seeds at recurrences 1--5, but only seed 43 has pairs at recurrence 6. Bands are 95\% relation-group bootstrap intervals.}
  \label{fig:frontier-answer}
\end{figure}

Figure~\ref{fig:frontier-answer} shows whether the model follows the patched entity until the end of the chain under different intervention methods. In particular, we intervene at each loop using the following methods:

\begin{itemize}
      \item \textbf{Full vector:} We replace the entire residual vector
      with the corresponding donor vector.
      \item \textbf{Entity subspace:} We transfer only the donor-base
      difference projected onto the entity subspace (as in Section~\ref{sec:causal_interventions}).
      \item \textbf{Aligned subspace:} We insert a validation-derived
      target-entity representation aligned across loops.
      \item \textbf{Residual-matched subspace:} We use a target-entity
      donor selected to minimize differences outside the entity subspace.
      \item \textbf{Random subspace:} We transfer an equal-dimensional,
      norm-matched random projection as a control.
      \item \textbf{Same-entity context:} For each base query, we construct untrained queries that share its relation suffix after the frontier but replace the relations before it by a different prefix, with a start entity chosen so that the chain reaches the same frontier entity. We keep those whose own frontier at that recurrence equals the selected one, draw one at random, and patch either its entity component or its whole state at the frontier position into the base. The answer should not change. Over recurrences 2--5 (2{,}221--2{,}304 pairs per recurrence; no such context is constructed at recurrence 1, where the frontier is 0), the base answer is preserved in 99.57--99.87\% of cases with the entity component and in 98.33--99.52\% with the whole state.
  \end{itemize}

\textbf{Frontier selection.} A bridge entity is \emph{ready} when the correct entity is ranked top-1 by its probe and its calibrated pointwise $\mathcal{V}$-information \citep{xu2020theoryusableinformationcomputational} exceeds a threshold $\beta$. For a probe with distribution $q$ over the $|\mathcal{E}|$ entities, the pointwise $\mathcal{V}$-information of the correct entity $e$ is $\mathrm{PVI}=\log_2 q(e\mid h)+\log_2|\mathcal{E}|$, the bits gained over a uniform guess. We set $\beta$ to the 10th percentile of PVI over correct validation decodes, so that $90\%$ of correct decodes are retained ($\beta\approx7.5$ bits in every seed). The frontier is the longest contiguous prefix of ready entities. For each recurrence we take the modal frontier on validation data: 0 at recurrence 1 (no bridge entity is yet ready), and 1, 3, 5 and 8 at recurrences 2--5. These hold for all three evaluation seeds except at recurrence 4, where one seed selects frontier 6 (816 vs.\ 719 of 1{,}600 validation chains for frontiers 6 and 5). Recurrence 6 does not provide a comparable three-seed intervention: seeds 41 and 42 select the terminal frontier 10 and yield no donor pairs, whereas seed 43 selects frontier 0 and yields 767 pairs, with no donor-implied final answers. We therefore omit it from the main comparison. A modal terminal frontier does not imply that most chains are complete. At recurrence 1, the random-subspace control reaches $50.95\%$ (CI $49.3$--$52.8\%$), so this recurrence is excluded from the comparison with the control.

\textbf{Projector construction.} The probes are affine classifiers of the $d=768$-dimensional LayerNorm-normalized state, trained with cross-entropy over the 217-token vocabulary. We construct the projector from their weight vectors but apply it to donor--base differences in the raw residual stream; we do not linearize or invert LayerNorm. There is one probe per loop, hop, and token position, read at the fourth layer of the shared block. For the intervention, we keep only the rows of the 200 entity classes, $W\in\mathbb{R}^{200\times768}$. We subtract the mean row, $\tilde W=W-\mathbf{1}\bar w^\top$. This subtracts the same scalar from every entity logit, preserving their relative logits and the entity-only softmax used for readiness. It need not preserve probabilities under the full 217-class softmax. We take the thin SVD of $\tilde W$ and keep the right singular vectors whose singular values exceed $10^{-5}$ times the largest. This gives an orthonormal basis $V\in\mathbb{R}^{768\times r}$ and the projector $P_{\mathrm{entity}}=VV^\top$. The centred rows sum to zero, so $r\le199$. In practice, $r=199$ at every intervention point and in every seed. The projector therefore spans $26\%$ of the hidden dimension and is not the identity. At recurrence $t\ge2$, we use the probe that decodes the frontier hop at the end of recurrence $t-1$, which is the same probe used to decide readiness. At recurrence 1, no bridge entity has been computed yet. The patched state is then the input embedding, so we build the same projector from the model's 200 entity token embeddings, centred in the same way (rank 199).

\textbf{Random control.} For each recurrence and seed, we draw a $768\times r$ Gaussian matrix with a fixed seed and orthonormalise it by a reduced QR decomposition, giving $Q$ with the same rank $r=199$ as the entity basis. We project the same donor--base difference $\Delta=h^{\mathrm{donor}}_p-h_p$ onto it and rescale it, pair by pair, to the norm of the entity-subspace patch: $h'_p=h_p+\|P_{\mathrm{entity}}\Delta\|_2\,QQ^\top\Delta/\|QQ^\top\Delta\|_2$. The control therefore matches the entity patch in both dimension and perturbation norm. After one recurrence it raises the descendants at offsets 1--2 by $3.2$--$8.4$ bits, against $15.1$--$18.0$ for the entity patch (Figure~\ref{fig:frontier-reach}). Each point of that figure uses 2{,}304 base--donor pairs over three seeds (2{,}286 at recurrence 4), averaging first within each relation group and then equally over the 72 seed-specific groups. Its 95\% CIs bootstrap these group means, with half-width at most $0.73$ bits at offsets 1--2, $1.03$ bits at offset 3, and $3.7$ percentage points for the top-1 rate.

\textbf{Pair selection and coverage.} Each seed uses 24 held-out relation groups of 200 ten-hop queries each, i.e.\ 4{,}800 test queries per recurrence. A query is eligible at recurrence $t$ when its observed frontier equals the frontier selected on validation data. Within each relation group, we shuffle the eligible queries and pair each one with a donor obtained by a cyclic shift of the shuffled list, so that base and donor are always distinct queries. We keep at most 32 pairs per group and recurrence. None of the retained pairs has the same frontier entity or the same final answer in base and donor. Table~\ref{tab:patch_counts} reports, for each recurrence, the selected frontier, the coverage, the number of retained pairs, and the final-answer accuracy of Figure~\ref{fig:frontier-answer}. Figures~\ref{fig:frontier-reach} and~\ref{fig:frontier-answer} use the same pairs. All results come from the single R6 checkpoint: seeds 41--43 vary the held-out relation groups and the donor selection, not the model initialisation. At recurrences 1--5, point estimates equally weight the 72 seed-specific relation-group means, and uncertainty is a 95\% percentile interval obtained by resampling those means.

\begin{table}[t]
\centering
\caption{Frontier patching on held-out ten-hop chains (R6, three seeds pooled). \emph{Frontier}: frontier selected on validation data (seeds 41/42/43). \emph{Coverage}: eligible test queries out of 14{,}400 (per-seed range in parentheses). \emph{Pairs}: retained base--donor pairs. The last three columns give final-answer accuracy (\%) for the donor-implied answer after six recurrences, equally averaged over relation groups, with 95\% group-bootstrap CIs.}
\label{tab:patch_counts}
\footnotesize
\setlength{\tabcolsep}{3.5pt}
\begin{tabular}{ccccccc}
\toprule
Rec. & Frontier & Coverage (\%) & Pairs & Entity subspace & Full vector & Random subspace \\
\midrule
1 & 0/0/0 & 100.0 (100.0--100.0) & 2{,}304 & 98.6 [98.1, 99.0] & 98.6 [98.1, 99.0] & 51.0 [49.3, 52.8] \\
2 & 1/1/1 & 96.8 (96.7--96.9) & 2{,}304 & 97.2 [96.5, 97.9] & 98.4 [97.8, 99.0] & 15.7 [14.1, 17.1] \\
3 & 3/3/3 & 91.5 (90.8--92.7) & 2{,}304 & 94.3 [93.1, 95.6] & 96.2 [95.1, 97.1] & 18.8 [17.1, 20.4] \\
4 & 6/5/5 & 51.4 (39.9--59.5) & 2{,}286 & 79.1 [76.8, 81.3] & 84.3 [82.4, 86.2] & 9.9 [8.7, 11.2] \\
5 & 8/8/8 & 67.0 (64.8--68.9) & 2{,}304 & 88.4 [86.5, 90.1] & 92.4 [90.8, 93.8] & 20.8 [18.9, 22.8] \\
\bottomrule
\end{tabular}
\end{table}

Entity-subspace patching, the main intervention, stays within $0$--$5.2$ percentage points of full-vector replacement (Table~\ref{tab:patch_counts}) and achieves at least 79\% counterfactual-answer accuracy at recurrences 1--5. The residual-matched variant at recurrences 2--5 stays within 0.8 points, although its available donor contexts produce slightly different sample counts (2{,}304, 2{,}304, 2{,}225 and 2{,}302). The aligned variant stays within 1.4 points except at recurrence 4, where it is 4.9 points higher (84.0\%, close to 84.3\% for full-vector replacement). The zero steering rate at recurrence 6 applies only to its selected seed-43 pairs and does not establish that computation has finished. Overall, the advantage over equal-dimensional, norm-matched random patches supports Q3: in this checkpoint and intervention setting, a subspace predictive of entity identity is also causally useful for steering the final answer.

\section{Further information on memorization across depths}\label{app:extra}

\begin{figure}[h]
\centering
\includegraphics[width=\linewidth]{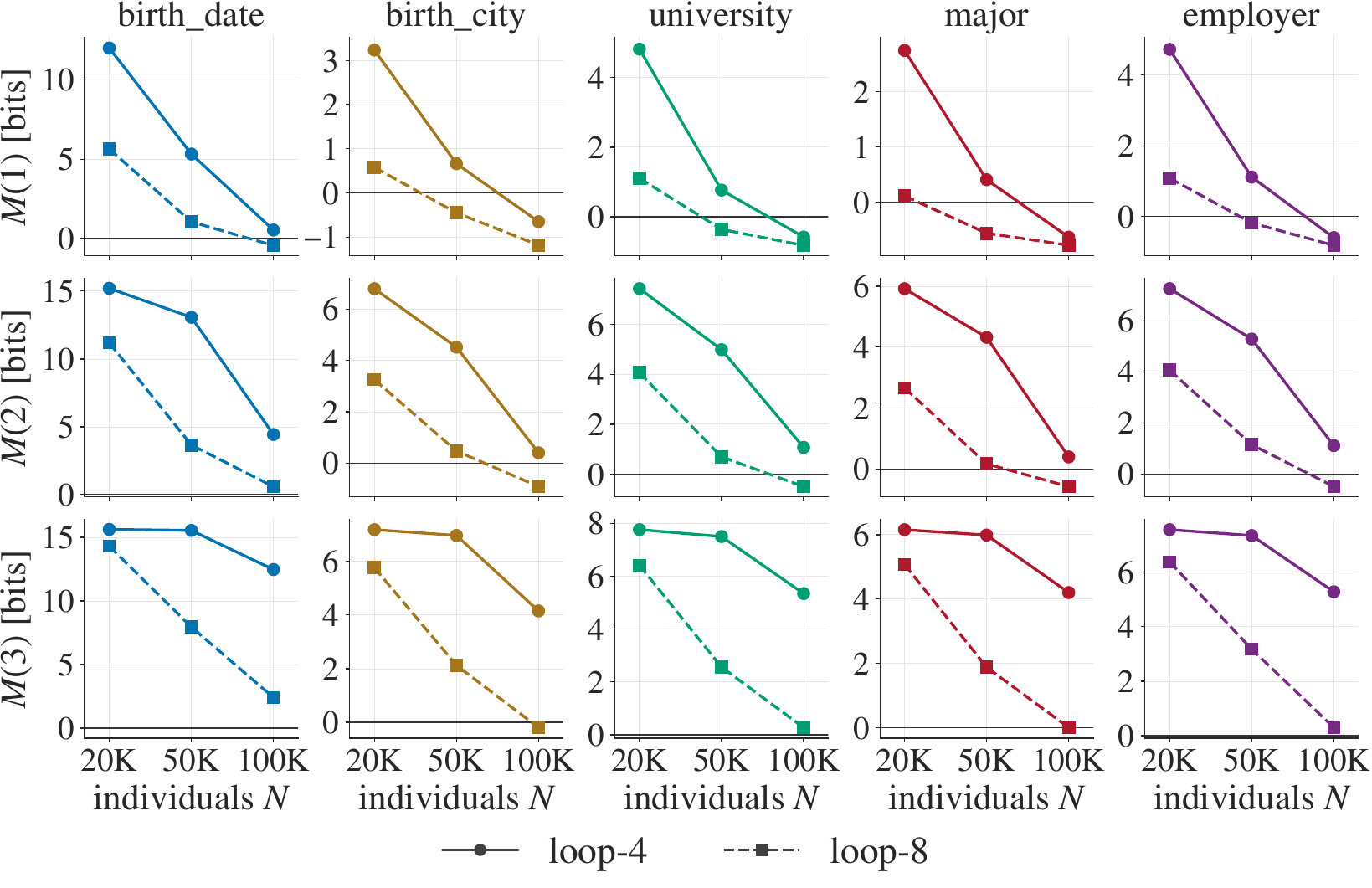}
\caption{\textbf{Early-depth readable bits, per attribute.}  Readable bits at $t = 1, 2, 3$ against dataset size, with each column a different attribute and rows
$t = 1, 2, 3$. Solid: loop-4; dashed: loop-8; thin line: $M = 0$.}
\label{fig:early_facet}
\end{figure}

\begin{figure}[!ht]
\centering
\includegraphics[width=\linewidth]{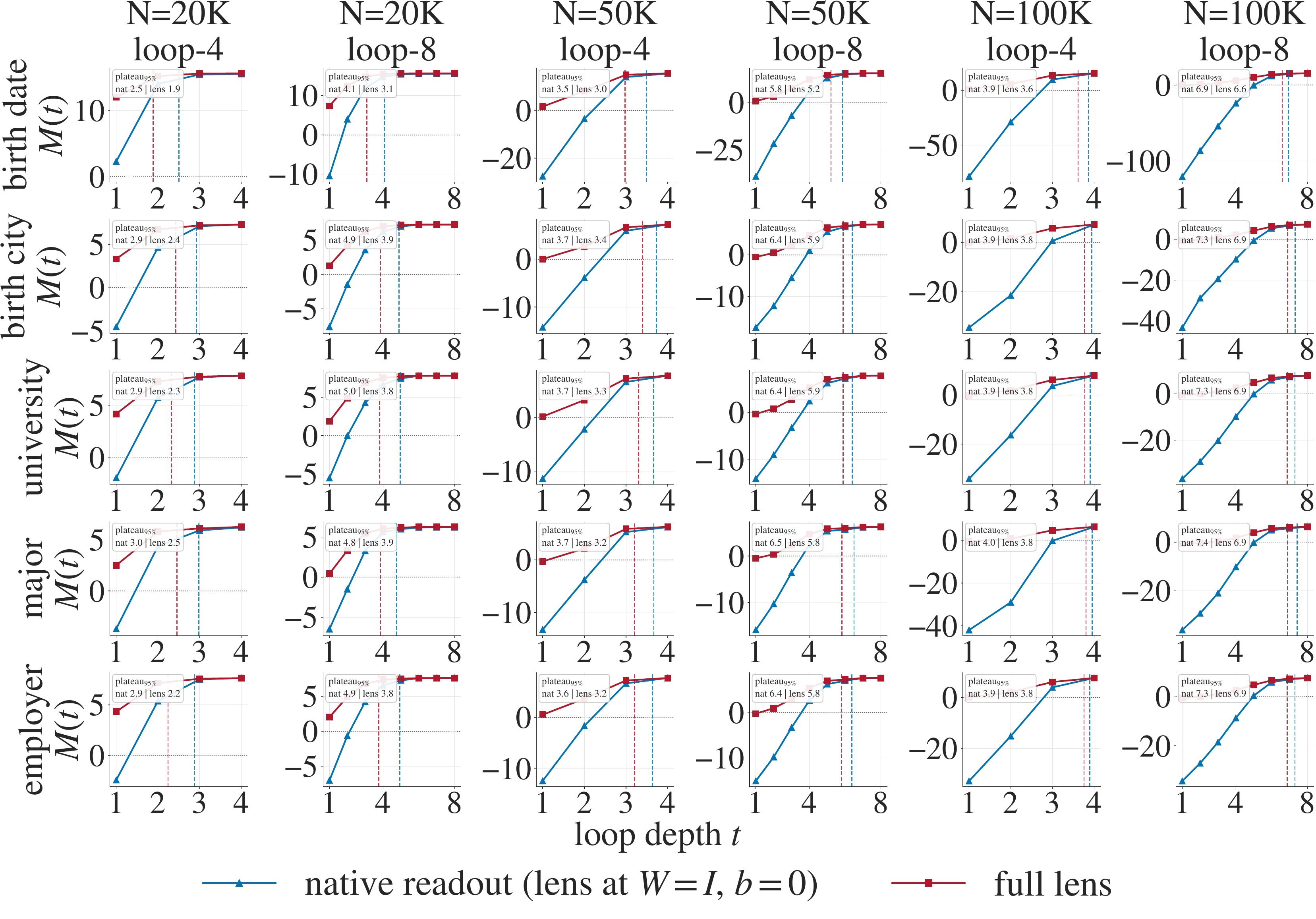}
\caption{\textbf{Native readout versus lens, per attribute.} As
Figure~\ref{fig:native_vs_lens}, one row per attribute. Plateau depths are
annotated per panel as ``nat $\mid$ lens'' (native readout $\mid$ full lens), and marked by dashed vertical lines in the matching colour. The dotted horizontal line marks $M = 0$. Seed 0.}
\label{fig:lens_facet}
\end{figure}

This appendix breaks the results of Figure~\ref{fig:native_vs_lens} down by attribute. Figure~\ref{fig:early_facet} shows how many bits are readable at the first three depths as the dataset grows, and Figure~\ref{fig:lens_facet} shows the full depth profile of every attribute under both readouts.

\textbf{Early-depth readable bits.} To identify where the load-dependent delay first appears, Figure~\ref{fig:early_facet} reports $M(1)$, $M(2)$ and $M(3)$, measured with the tuned lens, against the number of individuals $N$. These are seed-0 results for the 11-layer, $d=192$ block. Two regularities hold in all fifteen panels. First, readable bits at a fixed early depth decrease with $N$ for every attribute and both loop budgets. Second, at every $(t, N)$ point the loop-8 model has fewer readable bits than the loop-4 model. How large the gap between the two budgets is depends on $N$. At 20K it is smaller at $t=3$ than at $t=1$. At 50K and 100K it is instead larger at $t=3$ than at $t=1$, for every attribute.

At $t=1$ the decrease reaches the uniform-value log-loss baseline. With loop-4, $M(1)$ for the four single-token attributes drops from roughly $3$--$5$ bits at 20K to slightly below zero at 100K. Only birth date, whose $H_0$ is about twice as large, stays positive at 100K (about $0.6$ bits). With loop-8, $M(1)$ is already near zero at 20K for major and below zero at 50K for all four single-token attributes. At 100K it is negative for every attribute.

At $t=3$ the two budgets behave differently. With loop-4, $M(3)$ is almost flat from 20K to 50K (for birth date, $\approx 15.6$ and $\approx 15.5$ bits) and drops only at 100K. With loop-8, $M(3)$ falls steadily with $N$ and is close to zero at 100K for all single-token attributes. At 20K and 50K the loop-4 values at $t=3$ are already close to their final level (Figure~\ref{fig:lens_facet}), while the loop-8 values are still well below theirs. This matches the main-text observation that a larger loop budget spreads the same retrieval over more steps.

\textbf{Native readout versus lens, per attribute.} Figure~\ref{fig:lens_facet} shows $M(t)$ for the native readout and the full tuned lens, with one row per attribute and one column per $(N, L)$ cell. Each panel's ``plateau'' annotation denotes the linearly interpolated first crossing of $95\%$ of that curve's own final value, not a requirement that it remain above this threshold. All values are for seed 0. The per-attribute panels match the attribute-averaged picture in Figure~\ref{fig:native_vs_lens}:
\begin{itemize}
    \item 
\emph{Similar endpoints.} Native and lens curves end near the same value at $t=L$; equality is approximate, not imposed.
    \item \emph{The $95\%$ crossing moves later with $N$ for every attribute.} With the lens, crossing depths range over $1.9$--$2.3$ (20K), $2.7$--$3.0$ (50K) and $3.8$--$3.9$ (100K) for loop-4. For loop-8 the ranges are $3.6$--$4.7$, $5.1$--$5.9$ and $6.6$--$6.8$. The native readout follows the same ordering.
    \item \emph{The lens crosses earlier, but the gap shrinks with $N$.} In all $30$ panels the lens crossing depth is no larger than the native one. The difference is about $0.6$--$0.9$ loops at 20K, $0.4$--$0.6$ at 50K, and only $0.1$--$0.2$ at 100K. At the heaviest load, the two readouts therefore give similar near-final readability times.
    \item \emph{The native readout is strongly negative at shallow depths, more so as $N$ grows.} For the single-token attributes the native $M(1)$ lies between about $0$ and $-8$ bits at 20K, and between about $-29$ and $-47$ bits at 100K. For birth date it falls to about $-90$ (100K, loop-4) and $-130$ bits (100K, loop-8). At the same depths the lens is much closer to the baseline, although it too can be negative. The gap between the two readouts is largest at shallow depths and heavy loads.
    \item \emph{Birth date reaches its relative threshold first.} In every cell, birth date has the smallest or tied-smallest lens $95\%$ crossing depth among the five attributes.
\end{itemize}

\begin{figure}[t]
\centering
\includegraphics[width=0.6\linewidth]{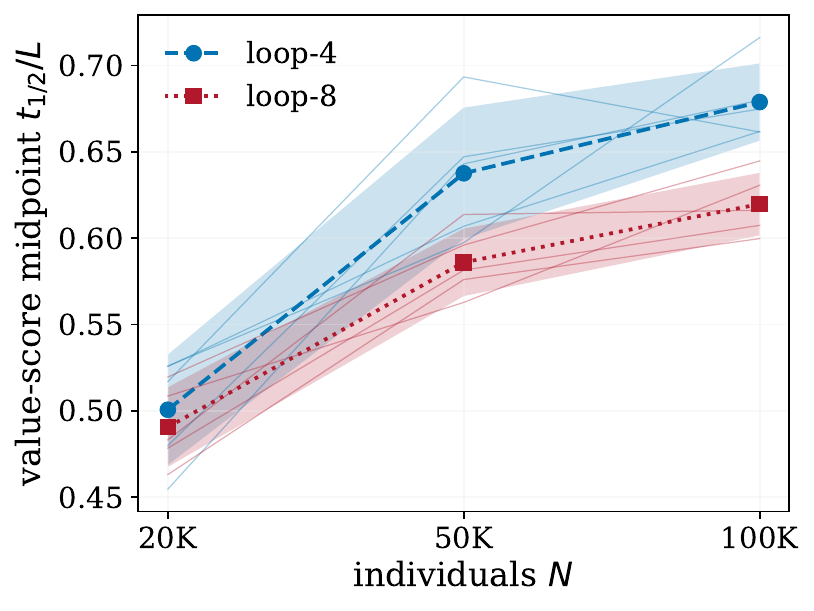}
\caption{
\textbf{Geometric midpoint} $t_{1/2}/L$ versus $N$: mean $\pm$1 SD over five seeds; thin lines: individual seeds. The midpoint uses excess permutation-normalized $F$, not $\Delta\eta^2$ (Appendix~\ref{app:keyvalue_timing}).}
\label{fig:seeds_thalf}
\end{figure}

\subsection{Timing of the key and value scores}\label{app:keyvalue_timing}
With more individuals, the key score of Figure~\ref{fig:key-value} stays at its initial level for longer: at loop-8 
it has fallen by 0.042, 0.027 and 0.011 at $t=6$, and by 0.12, 0.13 and 0.07 at $t=L$, for 20K, 50K and 100K. Never-trained individuals show a similar decline (0.14, 0.14 and 0.07 at $t=L$). Thus, a declining key score can occur without backbone training on those individuals' facts; it is not by itself evidence of fact retrieval. The value rise is likewise delayed: the midpoint $t_{1/2}$ moves later as $N$ grows (Figure~\ref{fig:seeds_thalf}), and for each single-token attribute its seed mean increases with $N$ at both loop budgets.

For each attribute and depth, let $R_a(t)=F_{{\rm value},a}(t)/\overline F_{{\rm perm},a}(t)$, where the denominator is that depth's mean permutation statistic. Set $E(t)=\frac14\sum_a R_a(t)-1$. We define $t_{1/2}$ by the first crossing of $E(L)/2$, interpolating linearly in $\log E$ between positive bracketing values; a crossing already attained at the first sampled loop is reported at that loop and is a censored upper bound on its depth. This is half the terminal excess, not half an inferred asymptotic plateau or half the plotted $\Delta\eta^2$. The mean paired-seed shifts from 20K to 50K are $2.4$ (loop-4) and $3.2$ (loop-8) times the sample SD of the paired differences; from 50K to 100K they are $0.8$ and $1.3$. These are descriptive standardized shifts, not significance tests.

\subsection{Untied models of equal depth}\label{app:untied}

For each $N$, we compare untied single-pass models with the looped models' unrolled depths, 44 and 88 layers (one run per cell, seed 0). Equal-width models keep $d=192$ and have 19.6M and 39.1M non-embedding parameters. Approximately parameter-matched models use $d=96$ at 44 layers (4.92M) and $d=66$ at 88 layers (4.68M), versus 4.89M in the looped models; the latter is about $4.5\%$ smaller. Data, A/B/C split, training steps, optimizer and global batch match; micro-batch/accumulation splits differ. Lenses are fit per layer. Final $M$ is $8.92$--$8.93$ bits except for the narrowed 88-layer, 100K model ($8.46$ bits). Looped models reach half-final $M$ at $0.23$--$0.65$ of unrolled depth, with later crossings at heavier loads. Untied crossings are $0.81$--$0.88$ (equal width) and $0.93$--$0.99$ (approximately matched parameters); their $95\%$ crossings are $0.94$--$1.00$ (Figure~\ref{fig:untied}, Table~\ref{tab:untied}). In narrowed models, the first positive $M$ moves from $0.57$ to $0.86$ of depth at 44 layers and $0.73$ to $0.94$ at 88 layers as load grows. Native $95\%$ crossings occur at $0.98$--$1.00$ in untied models versus $0.62$--$0.98$ in looped models.

\begin{table}[!ht]
\centering\small
\caption{
\textbf{Looped versus untied models of equal unrolled depth} (five-attribute mean on B; one run per cell, seed 0). Final full-lens $M$ and normalized first-crossing depths for 50\%/95\% of each readout's own final value, linearly interpolated. Looped models are sampled at loop boundaries; untied models at every layer. The 20K loop-4 half-final entry is already attained at the first sampled loop, so $0.25$ is an upper bound on its crossing depth. Parameter matching is approximate.}
\label{tab:untied}
\setlength{\tabcolsep}{3pt}
\begin{tabular}{ll ccc ccc ccc ccc}
\toprule
& & \multicolumn{3}{c}{final $M$ [bits]} & \multicolumn{3}{c}{lens, 50\%} & \multicolumn{3}{c}{lens, 95\%} & \multicolumn{3}{c}{native, 95\%} \\
\cmidrule(lr){3-5}\cmidrule(lr){6-8}\cmidrule(lr){9-11}\cmidrule(lr){12-14}
depth & model & 20K & 50K & 100K & 20K & 50K & 100K & 20K & 50K & 100K & 20K & 50K & 100K \\
\midrule
44 & looped (loop-4) & 8.93 & 8.93 & 8.93 & 0.25 & 0.40 & 0.65 & 0.50 & 0.73 & 0.96 & 0.71 & 0.87 & 0.98 \\
   & untied, equal width & 8.93 & 8.93 & 8.93 & 0.81 & 0.82 & 0.81 & 0.94 & 0.95 & 0.95 & 0.98 & 0.99 & 1.00 \\
   & untied, matched & 8.93 & 8.93 & 8.92 & 0.93 & 0.97 & 0.97 & 0.99 & 1.00 & 1.00 & 1.00 & 1.00 & 1.00 \\
\midrule
88 & looped (loop-8) & 8.93 & 8.93 & 8.93 & 0.23 & 0.42 & 0.61 & 0.51 & 0.69 & 0.84 & 0.62 & 0.75 & 0.86 \\
   & untied, equal width & 8.93 & 8.93 & 8.93 & 0.88 & 0.88 & 0.83 & 0.96 & 0.97 & 0.95 & 0.98 & 0.99 & 1.00 \\
   & untied, matched & 8.93 & 8.92 & 8.46 & 0.96 & 0.99 & 0.99 & 1.00 & 1.00 & 1.00 & 1.00 & 1.00 & 1.00 \\
\bottomrule
\end{tabular}
\end{table}

\end{document}